%% file: main.tex
\documentclass[letterpaper]{article} % DO NOT CHANGE THIS
\usepackage{aaai24}  % DO NOT CHANGE THIS
\usepackage{times}  % DO NOT CHANGE THIS
\usepackage{helvet}  % DO NOT CHANGE THIS
\usepackage{courier}  % DO NOT CHANGE THIS
\usepackage[hyphens]{url}  % DO NOT CHANGE THIS
\usepackage{graphicx} % DO NOT CHANGE THIS
\usepackage{natbib}  % DO NOT CHANGE THIS AND DO NOT ADD ANY OPTIONS TO IT
\usepackage{caption} % DO NOT CHANGE THIS AND DO NOT ADD ANY OPTIONS TO IT
\usepackage{algorithm}
\usepackage{algorithmic}

\usepackage{multirow}
\usepackage{amsmath}
\usepackage{mathtools,amssymb,lipsum}
\usepackage{commath}
\usepackage{pgfplots}
\pgfplotsset{compat=newest}
\usepackage{tikz}
\usepgfplotslibrary{groupplots}
\usetikzlibrary{patterns}
\usepackage{caption}
\usepackage{subcaption}
\usepackage{booktabs}
\usepackage{newfloat}
\usepackage{listings}
\DeclareCaptionStyle{ruled}{labelfont=normalfont,labelsep=colon,strut=off} % DO NOT CHANGE THIS
\floatstyle{ruled}
\newfloat{listing}{tb}{lst}{}
\floatname{listing}{Listing}
\title{Domain Adaptation for Time series Transformers using One-step fine-tuning}
\author{
   Subina Khanal\textsuperscript{\rm 1},
   Seshu Tirupathi\textsuperscript{\rm 2},
   Giulio Zizzo\textsuperscript{\rm 2},
   Ambrish Rawat\textsuperscript{\rm 2},
   Torben Bach Pedersen\textsuperscript{\rm 1}
}
\affiliations{
    \textsuperscript{\rm 1}Department of Computer Science, Aalborg University\\
    \textsuperscript{\rm 2}IBM Research Europe\\
    subinak@cs.aau.dk, seshutir@ie.ibm.com, giulio.zizzo2@ibm.com, ambrish.rawat@ie.ibm.com,
    tbp@cs.aau.dk
}

\usepackage{bibentry}
\begin{document}

\maketitle

\begin{abstract}
The recent breakthrough of Transformers in deep learning has drawn significant attention of the time series community due to their ability to capture long-range dependencies. However, like other deep learning models, Transformers face limitations in time series prediction, including insufficient temporal understanding, generalization challenges, and data shift issues for the domains with limited data. Additionally, addressing the issue of catastrophic forgetting, where models forget previously learned information when exposed to new data, is another critical aspect that requires attention in enhancing the robustness of Transformers for time series tasks. To address these limitations, in this paper, we pre-train the time series Transformer model on a source domain with sufficient data and fine-tune it on the target domain with limited data. We introduce the {\em One-step fine-tuning} approach, adding some percentage of source domain data to the target domains, providing the model with diverse time series instances. We then fine-tune the pre-trained model using a gradual unfreezing technique. This helps enhance the model's performance in time series prediction for domains with limited data. Extensive experimental results on two real-world datasets show that our approach improves over the state-of-the-art baselines by 4.35\% and 11.54\% for indoor temperature and wind power prediction, respectively.

\end{abstract}

\section{Introduction}
Time series prediction has been a significant subject of academic study with applications in finance, weather, and climate change. Time series prediction approaches have evolved over the past few decades from classical statistical methodologies and machine learning (ML) techniques to deep learning-based solutions. Recently, the breakthrough of Transformers in deep learning has attracted much attention from the time series community owing to its outstanding performance in a variety of computer vision and natural language processing tasks \cite{vaswani2017attention, wen2022transformers}. Transformers have many benefits, but one that makes time series modeling particularly well suited for them is their capacity to capture long-range dependencies and interactions. This has resulted in significant advancements in a variety of time series applications \cite{ahmed2023transformers}. However, Transformers, like other deep learning models such as Recurrent Neural Networks (RNNs), Convolutional Neural Networks (CNNs), or Autoencoders,  have several limitations on time series prediction. \emph{First}, \textbf{limited data availability:} deep learning models require a large amount of data for training, which is not always available for some time series prediction in real-world scenarios \cite{sun2017revisiting}. \emph{Second}, \textbf{lack of temporal understanding:} Transformers, in particular, may not accurately capture the temporal dynamics in time series because they are primarily intended for tasks not inherently involving temporal dependencies \cite{zeng2023transformers}. \emph{Third}, the \textbf{problem of data shift:} deep learning models assume that the training and test data are drawn from the same distribution. However, real-world scenarios often involve changes in data distribution, termed data shift. When this assumption is violated, the performance of the model degrades significantly \cite{farahani2021brief}. \emph{Fourth}, \textbf{lack of generalization:} deep learning models frequently have difficulty generalizing well to new, unseen data \cite{ying2019overview}. They may perform well on the training data but fail to accurately predict unseen data. This is the scenario where the unseen data is \textit{not independent and identically distributed (non-i.i.d)} w.r.t. the training data because of the data shift problem. Thus, the model becomes specialized to the training data and cannot perform well for all types of new data.

%This is the scenario of overfitting, where the model becomes specialized to the training data and cannot perform well for all types of new data.

\emph{Domain adaptation (DA)} is a technique used to address the above  mentioned challenges by improving the model's ability to generalize in scenarios with a different data distribution \cite{farahani2021brief, zhuang2020comprehensive}. DA leverages knowledge from the \emph{source domain}, where sufficient data is available, to the \emph{target domain}, where the data is limited, for making accurate predictions even when the two domains have different data distributions. However, models, like Transformers, are usually trained on fixed data, assuming the distribution of the test data remains constant. This is impractical for real-world applications, where data can evolve, and the model might need to adapt to new information without completely retraining from scratch \cite{kirkpatrick2017overcoming}. \emph{Continual learning} \cite{wang2023comprehensive} addresses this by allowing models to learn continuously from a data stream over time. The ML model can retain knowledge from previously learned tasks while learning and adapting continuously as new data becomes available. However, this causes catastrophic forgetting where performance on old tasks degrades over time as new tasks are added \cite{zizzo2022federated}. To address catastrophic forgetting in continual learning, replay methods \cite{bagus2021investigation} reintroduce past data during training that enables the model to balance learning from new tasks and retaining knowledge from earlier ones. It is also seen that catastrophic forgetting can occur during DA when a model is adapted to the target domain \cite{xu2020forget, saunders2022domain}. This is because of two reasons: \emph{First}, when performing DA, we train the prediction model on sufficient data from the source domain. Then, we leverage the knowledge of that model to the target domain with limited data to allow the model to generalize effectively on the target domain. However, during this adaptation process, the model may prioritize the data of the target domain and adjust its parameters in a way that causes it to forget the previously learned knowledge from the source domain. As a result, the performance of the model may degrade in the source domain, leading to generalization issues. \emph{Second}, there is a higher chance of forgetting if the data distributions for the source and target domains differ significantly. 

This paper aims to adapt the Transformer model to the target domain while simultaneously addressing the data shift and catastrophic forgetting problems between source and target domains. We pre-train a time series Transformer model on large data of the source domain and fine-tune and adapt the pre-trained model on different target domains using our \emph{One-step fine-tuning} approach. Therein, we specifically involve portion of source domain data to the target domains and fine-tune the pre-trained model using a gradual unfreezing (GU) \cite{howard2018universal} technique, which allows us to adapt and scale the pre-trained Transformer model to different target domains.

In summary, the main contributions of the present paper are as follows.
(1) We propose a \emph{One-step fine-tuning} technique, where we add some percentage of source domain data to the target domains and fine-tune the pre-trained model. Thereby, we show the fine-tuned model is adaptable and performs well on new, unseen data from different target domains. (2) We mitigate the problems of data shift and catastrophic forgetting by fine-tuning the pre-trained time series Transformer model on different target domains with limited data. (3) We conduct an extensive experimental evaluation using real-world datasets, which shows that the our \emph{One-step fine-tuning} approach outperforms the state-of-the-art baselines. We obtain 4.35\% and 11.54\% improvements over the most competitive baseline for indoor temperature and wind power prediction, respectively.

\section{Related Work}
Several Transformer-based time series forecasting techniques have gained significant attention recently, particularly for long-term time series forecasting. The authors in \cite{zhou2021informer} study the long-sequence time series forecasting problem and aim to predict long sequences. The ProbSparse self-attention mechanism and the distilling operation are used to handle the challenges of quadratic time complexity and quadratic memory usage in the vanilla Transformer. Also, this method alleviates the limitation of the traditional encoder-decoder architecture by using a carefully designed generative decoder. Similarly, in \cite{wu2021autoformer}, the authors study the problem of long-term forecasting of time series. They propose a decomposition architecture by embedding the series decomposition block as an inner operator, which can progressively aggregate the long-term trend part from intermediate prediction. This approach also designs an Auto-Correlation mechanism to conduct dependencies discovery and information aggregation at the series level, which differs significantly from the previous self-attention family. Likewise, in \cite{nie2022time}, the authors introduce two key components: 1) patching; segmentation of time series into subseries-level patches, which are served as input tokens to the Transformer; and 2) a channel-independent structure, where each channel contains a single univariate time series that shares the same embedding and Transformer weights across all the series for long-term multivariate time series forecasting and self-supervised representation learning.

\section{Preliminaries}
This section presents the core concepts of our \emph{One-step fine-tuning} approach for time series prediction that will be used throughout the paper.

\hspace{0.12in}\textbf{Time series:}
A \emph{univariate time series} $X = (x_{1}, x_{2},\ldots, x_n$), where $x_{t} \in \mathbb{R}$, is a sequence of real values that measure the same phenomenon and are chronologically ordered. Each value $x_t$ is recorded at uniformly spaced time instants $t\in \{1,2,\ldots\}$. A time series $X$ has length $n$ if it has $n$ collected samples.

\textbf{Time window of a time series:} 
Let $X$ be a time series of length $n$. A time window $X_w = (x_{t-m+1},\ldots, x_{t-1}, x_t)$ consists of historical values of $X$ recorded at time $(t-m+1)$ up to time $t$. Here, $m$ is the \emph{memory}, denoting the size of the time window $X_w$ which defines how many historical values will be used for prediction.

\textbf{Source domain:}  A \emph{source domain} $\mathrm{SD}$ is a set of training data with a given distribution of the input data $P(X_{\mathrm{SD}})$ used to train a prediction model. Here, we consider a training dataset with enough historical data, often collected over more than a year, as the source domain. 

\textbf{Target domain:} A \emph{target domain} $\mathrm{TD}$ is a set of data with a given  distribution of the input data $P(X_{\mathrm{TD}})$ that differs from the source domain distribution. Here, we consider target domains to have limited data.

\textbf{Learning task:} Both source and target domains have the same learning task $T$. Here, the learning task is to predict future time series using historical values. 

\textbf{Objective of the Source task:} Let us consider $X_{\mathrm{SD}}$ to be the input of the model $\mathcal{M}$, that maps $X_{\mathrm{SD}}$ to the predicted output $\hat{Y}_{\mathrm{SD}}$ in the source domain $\mathrm{SD}$.
\begin{equation}
   \hat{Y}_{\mathrm{SD}} = \mathcal{M}(X_{\mathrm{SD}};\theta), 
\end{equation}
where $\theta$ is the learnable parameters, i.e., weights of model $\mathcal{M}$. The objective during source task training is to minimize a source task loss function, denoted as $\mathcal{L}(\hat{Y}_{\mathrm{SD}}, Y_{\mathrm{SD}})$, where $Y_{\mathrm{SD}}$ is the actual output.
\begin{equation}
    \theta^* = \underset{\theta}{\arg\min}\frac{1}{N_{\mathrm{SD}}} \sum_{i=1}^{N_{\mathrm{SD}}} \mathcal{L}(\hat{Y}_{i}, Y_{i}),
\end{equation}
where $N_{\mathrm{SD}}$ is the total number of samples in source domain, $\hat{Y}_{i}$ is the predicted output for the $i^{\text{th}}$ sample, and ${Y}_{i}$ is the actual output for the $i^{\text{th}}$ sample in the source domain. 

\textbf{Fine-tuning Objective on the Target task:} Once pre-training of the source task is complete, the knowledge gained by the source model is transferred to the target task. During fine-tuning, we define a target loss function, denoted as $\mathcal{L}(\hat{Y}, Y)$, where $Y$ is the actual output.
\begin{equation}
    \theta_{\text{ft}} = \underset{\theta}{\arg\min}\frac{1}{N} \sum_{i=1}^{N} \mathcal{L}(\hat{Y}_{i}, Y_{i}; \theta^*),
\end{equation}
where $N$ is the total number of samples in target domain, $\hat{Y}_{i}$ is the predicted output for the $i^{\text{th}}$ sample, and $Y_{i}$ is the actual output for the $i^{\text{th}}$ sample in the target domain. Here, the goal is to update the model parameters to minimize the target loss $\mathcal{L}(\hat{Y}, Y)$, thereby improving the performance of the model on the target task in the target domain.

\section{Proposed Approach}
In this section, we present the detailed deployment design of our \emph{One-step fine-tuning} approach. We build on a  time series Transformer model, which is a type of neural network architecture specifically designed to process and model sequential data \cite{wen2022transformers}. The detailed workflow of the model architecture is as follows: 

\textbf{Training Dataset and Data Pre-processing:} Data collected from multiple residential buildings and wind turbines are used as the source and target training datasets. Before initiating the model training process, these training datasets are pre-processed to clean and format the data into input vectors. 

\textbf{Positional Encoding:} The first layer of the model architecture is positional encoding, which is used to add positional information to the input sequence vectors. In Transformers, positional encoding is essential for informing the model about the relative positions of the data points in the sequence \cite{vaswani2017attention}. For the model to comprehend the temporal relationships between data points, positional encoding is required because the timing and order of observations are critical in time series. The positional encoding is added to the input embeddings, where each input embedding corresponds to a data point in the time series. The positional encoding vector is determined based on each data point's position (timestamp) in the sequence.

We follow the sinusoidal positional encoding technique, initially introduced in the Transformer architecture for natural language processing tasks \cite{vaswani2017attention} defined as:
\begin{equation}
\text{PE}_{(pos, 2i)} = \sin\left(\frac{pos}{10000^{2i / d_{\text{model}}}} \right),
\end{equation}
\begin{equation}
\text{PE}_{(pos, 2i + 1)} = \cos\left(\frac{pos}{10000^{2i / d_{\text{model}}}} \right),
\end{equation}
where $pos$ is the position of time step in a sequence, $i$ is a index of the dimension, and $d_{\text{model}}$ is the embedding dimension.

\textbf{Encoder Layers:} The encoder layers are the second layer of the model architecture, responsible for processing the input sequence using self-attention mechanisms and feed-forward networks. These layers are stacked on each other to form the encoder \cite{zhou2021informer}. The initial sub-layer of an encoder is the Self-Attention Mechanism, which processes the input sequence vectors and allows the interaction between the values of input vectors. The mechanism learns to weigh the importance of each vector with respect to the others, capturing longer-range temporal dependencies and relationships in the time series. During training, the output of the self-attention part is added to the original inputs for gradient flow, and then layer normalization is applied. Each point in the sequence is individually processed by a feed-forward neural network, adding non-linearity to the encoder's transformations and enabling the learning of higher-level representations of the input data at each layer.

\textbf{Linear Decoder Layer:} The final layer is the linear decoder layer, a fully connected layer used to map the output of the encoder layer to the final prediction \cite{vaswani2017attention}. The input to a linear decoder layer consists of hidden representations, i.e., the output from the encoder layer. The fundamental operation of a linear decoder layer is a linear transformation represented by a fully connected (dense) neural network layer. The linear transformation projects the high-dimensional input representations into a lower-dimensional space, aligning the model's internal representation with the desired output space. Then, it is reshaped to match the target sequence length. 

Next, with the model architecture defined, we discuss the two phases of the workflow in our \emph{One-step fine-tuning} approach.

\hspace{0.12in}\textbf{1) Pre-training time series Transformer model in source domain:} In the first phase, we train a time series Transformer model, i.e., the \emph{source model}, using the data from the source domain until it converges to a certain level of prediction accuracy. The trained source model efficiently learns the temporal dependencies and patterns from the large and diverse data of the source domain that we transfer to the target domains with limited data. As the learning task is the same between the source and target domains, we validate the patterns learned by the source model benefits the target domains for such scenarios.

\textbf{2) Fine-tuning on Target Domains:} The second phase is fine-tuning the pre-trained time series Transformer model, i.e., the source model, on the target domains, allowing it to adapt to the target data. This domain adaptation process helps the model perform better in the target domains, even with limited data. For this, we apply our proposed {\em One-step fine-tuning} of the source model in the target domains to obtain a \emph{fine-tuned target model}. This approach includes the following steps:

(2a) We first add some percentage of randomly sampled source domain data to the target domains. Including time series instances of the source domain in the target domains jointly addresses three fundamental issues: (i) the problem of data scarcity, (ii) data distribution mismatch, as data from both domains gets involved during fine tuning, and (iii) catastrophic forgetting, as the source model retains useful data representations through data sharing, enabling better adaptation to the target domain without completely forgetting the knowledge gained from the source domain.

(2b) We then apply gradual unfreezing (GU) technique \cite{howard2018universal}, where each layer of source model is frozen at first and then gradually unfreezed during each training epoch, to obtain a fine-tuned target model. This allows keeping the source model knowledge intact, while the newly added layers of the model are made trainable, and subsequently, tailoring the model for the target domain. As such, this helps stabilize the training process and minimize catastrophic forgetting of pre-trained representations. We consider both source and target domains to have same learning task; hence, new layers are not added to the source model during fine-tuning.

The execution of GU steps for Energy Data (see details on \textbf{Datasets}) is as follows: (2b.1) In the first 10 epochs, the top layers, i.e., the decoder layers (output layer), are unfrozen, (2b.2) From epochs 10 to 20, we gradually unfreeze the layers closer to the input, i.e., we progressively unfreeze the subsequent layers in the Transformer encoder layer and train the model, and finally, (2b.3) after epochs 20 to 35, we unfreeze all the remaining layers of the source model and train the model. Here, we track the training loss over several epochs and implement early stopping to avoid overfitting. If the training loss increases or plateaus, we gradually unfreeze the layers. We fine-tune the source model until it converges to a certain level of prediction accuracy. 

\section{Experimental Evaluation}
In this section, we present the experimental evaluation of our \emph{One-step fine-tuning} approach for time series prediction in the source and target domains.

\subsection{Datasets}
We use the following real-world datasets to evaluate the performance of our \emph{One-step fine-tuning} approach:

\textbf{Energy Data:} We use two public energy datasets in our experiments. The first dataset comes from the New York State Energy Research and Development Authority (NYSERDA) \cite{NYSERDA}, which maintain data from residential buildings in New York State. These buildings are 100 to 600 square meters and have 50 geothermal heat pumps. The data includes observations on indoor temperature, outdoor temperature, and power consumption for about 12 months, with readings every 15 minutes.
     
The second dataset is collected from the Net-Zero Energy Residential Test Facility (NIST) \cite{nist}. This facility tests technologies for meeting residential energy needs with renewable energy. The data, collected over a year, simulates the energy usage of a family of four. The readings are taken every minute.
    
A 15-minute data granularity is selected to be used across all datasets, as it is usually used in electricity and flexibility markets. Thus, the NIST dataset is down-sampled by averaging power consumption readings to 15-minute intervals and taking the last indoor and outdoor temperature readings.

\textbf{MORE Data:} We use wind park data with 18 months of data from 11 wind turbines (WT2 to WT11) in a wind park provided by Engie as part of the MORE H2020 project \cite{more}. The dataset consists of SCADA data from the sensors on the wind turbines and weather data. Data included in this dataset are: \\
    - Power output: average, minimum, maximum, standard deviation over 10min. \\
    - Ambient temperature: average, minimum, maximum, standard deviation over 10min. \\
    - Wind speed: average, minimum, maximum, standard deviation over 10min.
    
There are over 840,000 samples from all the wind turbines with 1 hour data granularity. The target variable is minimum power output forecast for a 4 hour horizon at 1 hour intervals. 
\begin{table}[t!]
\centering
\caption{Parameters used in source model fine-tuning.}
\label{tab:param}
\begin{tabular}{|l|cc|}
\hline
\multicolumn{1}{|c|}{\multirow{2}{*}{Parameters}} & \multicolumn{2}{c|}{Values} \\ \cline{2-3} 
\multicolumn{1}{|c|}{} & \multicolumn{1}{l|}{Energy Data} & \multicolumn{1}{l|}{MORE Data} \\ \hline
No. of Input features & \multicolumn{1}{c|}{1} & 18 \\ \hline
Target Variable & \multicolumn{1}{c|}{\begin{tabular}[c]{@{}c@{}}Indoor\\ Temperature\end{tabular}} & \begin{tabular}[c]{@{}c@{}}Minimum Wind\\ Power\end{tabular} \\ \hline
Batch Size & \multicolumn{1}{c|}{8} & 24 \\ \hline
Training Epoch & \multicolumn{1}{c|}{35} & 20 \\ \hline
Historical values & \multicolumn{1}{l|}{96 (i.e. 24 hours)} & 24 (i.e. 24 hours) \\ \hline
Prediction horizon & \multicolumn{1}{l|}{4 (i.e. 1 hour)} & 4 (i.e. 4 hours) \\ \hline
Learning rate & \multicolumn{1}{c|}{0.001} & 0.00001 \\ \hline
Training Set: Test Set & \multicolumn{2}{c|}{70\%:30\%} \\ \hline
Optimizer & \multicolumn{2}{c|}{Adam} \\ \hline
Model Loss Function & \multicolumn{2}{c|}{Mean Absolute Error (MAE)} \\ \hline
\end{tabular}
\end{table}
\begin{table}[t!]
\caption{Prediction error of source model compared to the model baselines.}
\label{tab:ptm}
\centering
\begin{tabular}{|l|ll|ll|}
\hline
\multirow{3}{*}{Methods} & \multicolumn{2}{l|}{Energy Data} & \multicolumn{2}{l|}{MORE Data} \\ \cline{2-5} 
 & \multicolumn{2}{l|}{\begin{tabular}[c]{@{}l@{}}Prediction Error \\ in S4\end{tabular}} & \multicolumn{2}{l|}{\begin{tabular}[c]{@{}l@{}}Prediction Error \\ in WT11\end{tabular}} \\ \cline{2-5} 
 & \multicolumn{1}{l|}{RMSE} & MAE & \multicolumn{1}{l|}{RMSE} & MAE \\ \hline
\textbf{Informer} & \multicolumn{1}{l|}{0.420} & 0.293 & \multicolumn{1}{l|}{5.389} & 3.251 \\ \hline
\textbf{Autoformer} & \multicolumn{1}{l|}{0.402} & 0.307 & \multicolumn{1}{l|}{4.830} & 3.198 \\ \hline
\textbf{PatchTST} & \multicolumn{1}{l|}{0.383} & 0.296 & \multicolumn{1}{l|}{4.417} & 2.334 \\ \hline
\textbf{LSTM} & \multicolumn{1}{l|}{0.243} & 0.185 & \multicolumn{1}{l|}{0.136} & 0.109 \\ \hline
\textbf{Linear Regression} & \multicolumn{1}{l|}{0.264} & 0.208 & \multicolumn{1}{l|}{0.139} & 0.116 \\ \hline
\textbf{Source Model} & \multicolumn{1}{l|}{\textbf{0.236}} & \textbf{0.180} & \multicolumn{1}{l|}{\textbf{0.118}} & \textbf{0.075} \\ \hline
\end{tabular}
\end{table}
\input{sdavsp}
\input{baselines}
\input{eavsp}
\input{mavsp}
\input{cf}

\subsection{Implementation Details}
We use PyTorch framework for the model implementation and evaluation. The implementation starts by selecting the source and target domains as follows.

\textbf{Selection of source domain for pre-training:} The source domain should have adequate training data with seasonal variations. Therefore, for Energy data, out of $50$ geothermal heat pump sites from the NYSERDA dataset, we have selected deployment site \emph{S4} as the source domain. Similarly, for MORE data, out of $11$ wind turbines, we have selected \emph{WT11} as the source domain.

\textbf{Selection of target domains for fine-tuning:} The target domains should have the higher data dispersion compared to the selected source domain, which leads to better generalization capability of a model, as it can capture the specific patterns and characteristics unique to the target domain. To compute the marginal distribution disparity between the source and target domains, we adopt a widely used loss function, Maximum Mean Discrepancy (MMD) \cite{gretton2012optimal, khanal2023fda}. MMD measures the non-parametric distances between the source and target domains by converting them into a Reproducing Kernel Hilbert Space (RKHS), computed as
\begin{align}
\begin{split}
        {}\mathrm{MMD}[P(X_{\mathrm{SD}}),P(X_{\mathrm{TD}})] = \norm{\mathbb{E}[\varphi(X_{\mathrm{SD}})]- \mathbb{E}[\varphi(X_{\mathrm{TD}})]}_{\mathcal{H}}^2 \\
        = \norm{\frac{1}{N_{\mathrm{SD}}}\sum\nolimits_{p=1}^{N_{\mathrm{SD}}}\varphi(x_p) - \frac{1}{N}\sum\nolimits_{q=1}^{N}\varphi(x_q)}_{\mathcal{H}}^2,
        \label{mmd}
    \end{split}
\end{align}
where $P(X_{\mathrm{SD}})$ and $P(X_{\mathrm{TD}})$ are the marginal data distributions of the source and target domains, respectively, $N_{\mathrm{SD}}$ and $N$ are the number of data samples in the source and target domains, $\varphi(\cdot)$ is the mapping function from the original feature space to the RKHS, and $\mathcal{H}$ is the  RHKS space. The value of MMD starts from $0$, signifying that the compared domains are the same.  

The target domains are selected based on MMD values to cover the distribution range of all the target domains. For Energy data, we used the Net-Zero dataset (NIST) and four different deployment sites (S5, S8, S15, and S49) from NYSERDA as the target domains to fine-tune the source model, with MMD values ranging from 0.089 to 0.133. Likewise, for MORE data, we used WT4, WT6, WT7, WT8, and WT10 as the target domains to fine-tune the source model, with MMD values ranging from 0.095 to 0.100.
\begin{table*}[t!]
\caption{Data Shift check on other target domains.}
\centering
\label{tab:ds}
\begin{tabular}{|c|c|cccccccc|llllllll}
\cline{1-10}
\multirow{10}{*}{Energy Data} & \multirow{3}{*}{Domains} & \multicolumn{8}{c|}{Prediction Error on Other Target Domains} &  &  &  &  &  &  &  &  \\ \cline{3-10}
 &  & \multicolumn{2}{c|}{S3} & \multicolumn{2}{c|}{S7} & \multicolumn{2}{c|}{S19} & \multicolumn{2}{c|}{S50} &  &  &  &  &  &  &  &  \\ \cline{3-10}
 &  & \multicolumn{1}{c|}{RMSE} & \multicolumn{1}{c|}{MAE} & \multicolumn{1}{c|}{RMSE} & \multicolumn{1}{c|}{MAE} & \multicolumn{1}{c|}{RMSE} & \multicolumn{1}{c|}{MAE} & \multicolumn{1}{c|}{RMSE} & MAE &  &  &  &  &  &  &  &  \\ \cline{2-10}
 & \begin{tabular}[c]{@{}c@{}}Source Domain (S4)\end{tabular} & \multicolumn{1}{c|}{0.378} & \multicolumn{1}{c|}{0.320} & \multicolumn{1}{c|}{1.09} & \multicolumn{1}{c|}{0.892} & \multicolumn{1}{c|}{0.320} & \multicolumn{1}{c|}{0.260} & \multicolumn{1}{c|}{0.594} & 0.495 &  &  &  &  &  &  &  &  \\ \cline{2-10}
 & \begin{tabular}[c]{@{}c@{}}Target Domains\end{tabular} & \multicolumn{8}{c|}{} &  &  &  &  &  &  &  &  \\ \cline{2-10}
  & S5 & \multicolumn{1}{c|}{0.222} & \multicolumn{1}{c|}{0.161} & \multicolumn{1}{c|}{0.345} & \multicolumn{1}{c|}{0.276} & \multicolumn{1}{c|}{0.249} & \multicolumn{1}{c|}{0.188} & \multicolumn{1}{c|}{0.383} & 0.294 \\ \cline{2-10} 
 & S8 & \multicolumn{1}{c|}{0.264} & \multicolumn{1}{c|}{0.202} & \multicolumn{1}{c|}{0.403} & \multicolumn{1}{c|}{0.328} & \multicolumn{1}{c|}{0.227} & \multicolumn{1}{c|}{0.169} & \multicolumn{1}{c|}{0.397} & 0.302 \\ \cline{2-10} 
 & S15 & \multicolumn{1}{c|}{0.221} & \multicolumn{1}{c|}{0.165} & \multicolumn{1}{c|}{0.334} & \multicolumn{1}{c|}{0.259} & \multicolumn{1}{c|}{0.251} & \multicolumn{1}{c|}{0.186} & \multicolumn{1}{c|}{0.365} & 0.279 \\ \cline{2-10} 
 & S49 & \multicolumn{1}{c|}{0.247} & \multicolumn{1}{c|}{0.197} & \multicolumn{1}{c|}{0.349} & \multicolumn{1}{c|}{0.265} & \multicolumn{1}{c|}{0.233} & \multicolumn{1}{c|}{0.178} & \multicolumn{1}{c|}{0.442} & 0.339 \\ \cline{2-10} 
 & NIST & \multicolumn{1}{c|}{0.315} & \multicolumn{1}{c|}{0.239} & \multicolumn{1}{c|}{0.376} & \multicolumn{1}{c|}{0.282} & \multicolumn{1}{c|}{0.232} & \multicolumn{1}{c|}{0.171} & \multicolumn{1}{c|}{0.409} & 0.321 &  &  &  &  &  &  &  &  \\ \cline{1-10}
\multirow{10}{*}{MORE Data} & \multirow{3}{*}{Domains} & \multicolumn{8}{c|}{Prediction Error on Other Target Domains} & \multicolumn{1}{c}{} & \multicolumn{1}{c}{} &  &  &  &  &  &  \\ \cline{3-10}
 &  & \multicolumn{2}{c|}{WT2} & \multicolumn{2}{c|}{WT3} & \multicolumn{2}{c|}{WT5} & \multicolumn{2}{c|}{WT9} & \multicolumn{2}{c}{} & \multicolumn{1}{c}{} & \multicolumn{2}{c}{} & \multicolumn{1}{c}{} &  &  \\ \cline{3-10}
 &  & \multicolumn{1}{c|}{RMSE} & \multicolumn{1}{c|}{MAE} & \multicolumn{1}{c|}{RMSE} & \multicolumn{1}{c|}{MAE} & \multicolumn{1}{c|}{RMSE} & \multicolumn{1}{c|}{MAE} & \multicolumn{1}{c|}{RMSE} & MAE & \multicolumn{1}{c}{} & \multicolumn{1}{c}{} &  &  &  &  &  &  \\ \cline{2-10}
 & \begin{tabular}[c]{@{}c@{}}Source Domain (WT11)\end{tabular} & \multicolumn{1}{c|}{0.133} & \multicolumn{1}{c|}{0.096} & \multicolumn{1}{c|}{0.160} & \multicolumn{1}{c|}{0.118} & \multicolumn{1}{c|}{0.151} & \multicolumn{1}{c|}{0.113} & \multicolumn{1}{c|}{0.152} & 0.112 & \multicolumn{1}{c}{} &  &  &  &  &  &  &  \\ \cline{2-10}
 & \begin{tabular}[c]{@{}c@{}}Target Domains\end{tabular} & \multicolumn{8}{c|}{} &  &  &  &  &  &  &  &  \\ \cline{2-10}
 & WT4 & \multicolumn{1}{c|}{0.121} & \multicolumn{1}{c|}{0.080} & \multicolumn{1}{c|}{0.132} & \multicolumn{1}{c|}{0.087} & \multicolumn{1}{c|}{0.115} & \multicolumn{1}{c|}{0.076} & \multicolumn{1}{c|}{0.121} & 0.078 & \multicolumn{1}{c}{} &  &  &  &  &  &  &  \\ \cline{2-10}
 & WT6 & \multicolumn{1}{c|}{0.120} & \multicolumn{1}{c|}{0.080} & \multicolumn{1}{c|}{0.132} & \multicolumn{1}{c|}{0.085} & \multicolumn{1}{c|}{0.114} & \multicolumn{1}{c|}{0.075} & \multicolumn{1}{c|}{0.120} & 0.078 & \multicolumn{1}{c}{} &  &  &  &  &  &  &  \\ \cline{2-10}
 & WT7 & \multicolumn{1}{c|}{0.118} & \multicolumn{1}{c|}{0.078} & \multicolumn{1}{c|}{0.128} & \multicolumn{1}{c|}{0.084} & \multicolumn{1}{c|}{0.116} & \multicolumn{1}{c|}{0.077} & \multicolumn{1}{c|}{0.119} & 0.077 & \multicolumn{1}{c}{} &  &  &  &  &  &  &  \\ \cline{2-10}
 & WT8 & \multicolumn{1}{c|}{0.124} & \multicolumn{1}{c|}{0.081} & \multicolumn{1}{c|}{0.138} & \multicolumn{1}{c|}{0.089} & \multicolumn{1}{c|}{0.119} & \multicolumn{1}{c|}{0.076} & \multicolumn{1}{c|}{0.126} & 0.081 & \multicolumn{1}{c}{} &  &  &  &  &  &  &  \\ \cline{2-10}
 & WT10 & \multicolumn{1}{c|}{0.131} & \multicolumn{1}{c|}{0.087} & \multicolumn{1}{c|}{0.130} & \multicolumn{1}{c|}{0.086} & \multicolumn{1}{c|}{0.118} & \multicolumn{1}{c|}{0.077} & \multicolumn{1}{c|}{0.122} & 0.078 & \multicolumn{1}{c}{} &  &  &  &  &  &  &  \\ \cline{1-10}
\end{tabular}
\end{table*}

\textbf{Source model fine-tuning parameters:} Table~\ref{tab:param} shows the summary of parameters used during source model fine-tuning in the selected target domains for Energy and MORE data. The hyper-parameters are obtained through grid search.

\subsection{Baselines}
\hspace{0.12in}\textbf{1. Model Baselines:} We have selected five different model baselines to evaluate our source model; including \textbf{Informer} \cite{zhou2021informer}, \textbf{Autoformer} \cite{wu2021autoformer}, \textbf{PatchTST }\cite{nie2022time}, \textbf{LSTM }\cite{staudemeyer2019understanding}, and \textbf{Linear Regression} \cite{liu2021forecast,james2023linear}.

\textbf{2. Fine-tuning Baselines:} We use the following fine-tuning baselines to evaluate our \emph{One-step fine-tuning} approach. \textbf{Gradual unfreezing (GU)} \cite{howard2018universal}: This technique gradually unfreezes the layers during fine-tuning rather than unfreezing all the layers at once, without including data from the source domain.
\textbf{Elastic Weight Consolidation (EWC)} \cite{barone2017regularization, kirkpatrick2017overcoming}: EWC is a regularization technique used to mitigate catastrophic forgetting in neural networks, particularly in the context of continual learning. This technique adds penalty terms or constraints to the model training process. It modifies the model loss/objective function by adding a regularization term.
    
\textbf{3. Model Training Baselines:} We also compare the performance of our \emph{One-step fine-tuning} approach with other model training baselines as follows. \textbf{Exclusive Training:} Here, we train the time series Transformer model individually in different target domains and evaluate the individual target models on a test set of the target domain data. \textbf{Before fine-tuning:} Here, we directly assess the source model in the target domains without fine-tuning. 

\subsection{Time Series Prediction Evaluation}
\hspace{0.12in}\textbf{Evaluation Metrics:}
We use well-known metrics, Root Mean Square Error (RMSE) and Mean Absolute Error (MAE), for the quantitative assessment of source and final global model performance on test target domains. These metrics are calculated as
\begin{equation}
    \mathit{RMSE}(Y_{i}, \hat{Y}_{i}) = \sqrt{\frac{1}{N}\sum_{i=1}^{N}(Y_{i} - \hat{Y}_{i})^2},
\end{equation}
\begin{equation}
   \mathit{MAE}(Y_{i}, \hat{Y}_{i}) = \frac{1}{N}\sum_{i=1}^{N}\abs{(Y_{i} - \hat{Y}_{i})},
\end{equation}
where $Y_{i}$ and $\hat{Y}_{i}$ are the actual and predicted output, and $N$ is the total number of samples in the target domains.

\textbf{Prediction Accuracy Evaluation:} Table~\ref{tab:ptm} shows the quantitative performance of the source model evaluated on the source domains S4 and WT11 based on RMSE and MAE metrics respectively. Here, we observe a significant improvement in the source model's performance against the baselines for S4 and WT11, with up to 44\% and 97.8\% RMSE error reduction, respectively. Likewise, in Fig~\ref{fig:avsp_sd}, we have the prediction performance of the source model on the source domains S4 and WT11, respectively. We can see that the source model obtains improved prediction of the indoor temperature and wind power for both source domains. 

Next, Fig~\ref{fig:ftm} shows the performance of the fine-tuned target model. Our \emph{One-step fine-tuning} approach performs better than the baselines \textbf{GU} and \textbf{EWC} on all target domains with limited target data, with a significant improvement of up to 48.5\% after adding a portion of source domain data prior to fine-tuning the source model. For Energy data, we add 5\% (2731 samples) of source domain (S4) data on the data of S5, S49 and NIST as the MMD values between these target domains with S4 are higher. We add 20\% (10927 samples) of S4 data on S8 and S15 as the MMD values between these target domains with S4 are lower (see details on \textbf{Appendix-Ablation Study}). Similarly, for MORE data, we add 5\% (6796 samples) of source domain (WT11) data on the data of all the target domains. In Fig~\ref{fig:eba}, we show the comparison between different model training baselines: \textbf{Exclusive Training} and \textbf{Before fine-tuning}, with our \emph{One-step fine-tuning} approach. Here, we observe that the fine-tuned target model performs up to 52.6\% better than the baselines. This is because the fine-tuned target model can generalize well to new, unseen data through the learned general features and patterns from the source model, mitigating the problem of data shift. The baselines do not consider the data shift problem and perform well only if the test data follows the distribution of the training data. 

In Fig~\ref{fig:energy_avsp_td} and Fig~\ref{fig:more_avsp_td}, we illustrate the prediction performance of the fine-tuned target models on the target domains NIST and WT8, respectively. Here, we compare our \emph{One-step fine-tuning} approach with other fine-tuning baselines, where we can see that the fine-tuned target models obtains improved prediction of the indoor temperature and wind power for both target domains as compared to the baselines. 

In Table~\ref{tab:ds}, we show evaluation results for our \emph{One-step fine-tuning} approach under data shift scenario. This is done by directly applying both the pre-trained (source) and fine-tuned target models on other new target domains, i.e., without exclusive training and fine-tuning, respectively. We observe that the performance of the fine-tuned target model is better than the source model for all the other target domains, even though having different MMD values (see Table~\ref{tab:mmd} and Table~\ref{tab:mmd_more} in \textbf{Appendix}) that indicates data distribution scenario. This validates our claim to efficiently handle data shift problem; the fine-tuned target model performs better on unseen data by improving generalization, leading to better domain adaptation. 

Next, in Fig~\ref{fig:cf}, we show the performance of the fine-tuned target models directly evaluated on the source domains S4 and WT11 for catastrophic forgetting check. Here, we observe the fine-tuned target models using our \emph{One-step fine-tuning} approach outperforms on both S4 and WT11 than the fine-tuning baselines \textbf{GU} and \textbf{EWC}. The \emph{One-step fine-tuning} method achieves RMSE that is better or closer to the performance of the source model on S4, i.e., $0.236$. We obtain similar results for WT11, where the RMSE was $0.118$. This shows the fine-tuned model retains its knowledge learned during pre-training while adapting better to the new target domain data; hence, mitigating the problem of catastrophic forgetting by adding some percentage of data samples of S4 and WT11 to the respective target domains during model training.

\section{Conclusion}
Transformers improve time series prediction by effectively capturing long-term patterns and dependencies, improving the model's capability to understand and predict complex temporal relationships. However, their performance suffers severely due to limitations like insufficient temporal understanding, generalization challenges, data shift issues, and the critical concern of catastrophic forgetting. Therefore, in this paper, we propose a new \emph{One-step fine-tuning} approach where we pre-train the time series Transformer model on a source domain with sufficient data and fine-tune it on the target domain with limited data. Our approach builds on a gradual unfreezing technique, however, with addition of source domain data in the target domains to fine-tune the pre-trained model. This helps enhance the model's performance in time series prediction for domains with limited data. In the future, we will explore privacy aspects of our \emph{One-step fine-tuning} when sharing source domain data.

\section{Acknowledgments}
This work has been partially supported by the MORE project (grant agreement 957345), funded by the EU Horizon 2020 program.

\bibliography{aaai24}

\section{Appendix}
\begin{table*}[t!]
\caption{Overview of dataset used during pre-training and \emph{One-step fine-tuning}.}
\centering
\label{tab:dataset}
\begin{tabular}{|ccc|ccc|}
\hline
\multicolumn{3}{|c|}{Energy Data} & \multicolumn{3}{c|}{MORE Data} \\ \hline
\multicolumn{1}{|c|}{Domain} & \multicolumn{1}{c|}{Training Data} & Test Data & \multicolumn{1}{c|}{Domain} & \multicolumn{1}{c|}{Training Data} & Test Data \\ \hline
\multicolumn{1}{|c|}{Source Domain (S4)} & \multicolumn{1}{c|}{38,245} & 16,390 & \multicolumn{1}{c|}{Source Domain (WT11)} & \multicolumn{1}{c|}{95,148} & 40,777 \\ \hline
\multicolumn{1}{|c|}{S5} & \multicolumn{1}{c|}{4097} & 1755 & \multicolumn{1}{c|}{WT4} & \multicolumn{1}{c|}{10,850} & 4650 \\ \hline
\multicolumn{1}{|c|}{S8} & \multicolumn{1}{c|}{10,070} & 4316 & \multicolumn{1}{c|}{WT6} & \multicolumn{1}{c|}{10,794} & 2587 \\ \hline
\multicolumn{1}{|c|}{S15} & \multicolumn{1}{c|}{8830} & 3784 & \multicolumn{1}{c|}{WT7} & \multicolumn{1}{c|}{9994} & 2569 \\ \hline
\multicolumn{1}{|c|}{S49} & \multicolumn{1}{c|}{5700} & 2442 & \multicolumn{1}{c|}{WT8} & \multicolumn{1}{c|}{8128} & 2626 \\ \hline
\multicolumn{1}{|c|}{NIST} & \multicolumn{1}{c|}{5439} & 2331 & \multicolumn{1}{c|}{WT9} & \multicolumn{1}{c|}{9014} & 2577 \\ \hline
\end{tabular}
\end{table*}
\begin{table}[t!]
\caption{Parameters used in source model pre-training.}
\centering
\label{tab:param_source}
\begin{tabular}{|c|cc|}
\hline
\multirow{2}{*}{Parameters} & \multicolumn{2}{c|}{Values} \\ \cline{2-3} 
 & \multicolumn{1}{c|}{Energy Data} & MORE Data \\ \hline
No. of Input features & \multicolumn{1}{c|}{1} & 18 \\ \hline
Target Variable & \multicolumn{1}{c|}{\begin{tabular}[c]{@{}c@{}}Indoor\\ Temperature\end{tabular}} & \begin{tabular}[c]{@{}c@{}}Minimum Wind\\  Power\end{tabular} \\ \hline
Batch Size & \multicolumn{1}{c|}{8} & 64 \\ \hline
Training Epoch & \multicolumn{1}{c|}{50} & 100 \\ \hline
Historical values & \multicolumn{1}{c|}{96 (i.e. 24 hours)} & 24 (i.e. 24 hours) \\ \hline
Prediction horizon & \multicolumn{1}{c|}{4 (i.e. 1 hour)} & 4 (i.e. 4 hours) \\ \hline
Learning rate & \multicolumn{1}{c|}{0.001} & 0.0001 \\ \hline
Training Set: Test Set & \multicolumn{2}{c|}{70\%:30\%} \\ \hline
Optimizer & \multicolumn{2}{c|}{Adam} \\ \hline
Model Loss Function & \multicolumn{2}{c|}{Mean Absolute Error (MAE)} \\ \hline
\end{tabular}
\end{table}
\input{vp}
\input{ft}
\begin{table}[t!]
\caption{Calculation of MMD values between source and target domains for Energy data.}
\centering
\label{tab:mmd1}
\begin{tabular}{|c|c|c|}
\hline
Source Domain & Target Domain & MMD values \\ \hline
\multirow{5}{*}{S4} & S5 & 0.114 \\ \cline{2-3} 
 & S8 & 0.097 \\ \cline{2-3} 
 & S15 & 0.089 \\ \cline{2-3} 
 & S49 & 0.104 \\ \cline{2-3} 
 & NIST & 0.133 \\ \hline
\end{tabular}
\end{table}
\begin{table*}[t!]
\caption{Calculation of MMD values between source and target domains for Energy data.}
\centering
\label{tab:mmd}
\begin{tabular}{|c|c|c|c|c|c|}
\hline
Source Domain       & Other Target Domains & MMD values & Target Domain         & Other Target Domains & MMD values \\ \hline
\multirow{4}{*}{S4} & S3                   & 0.127      & \multirow{4}{*}{S15}  & S3                   & 0.058      \\ \cline{2-3} \cline{5-6} 
                    & S7                   & 0.153      &                       & S7                   & 0.157      \\ \cline{2-3} \cline{5-6} 
                    & S19                  & 0.131      &                       & S19                  & 0.107      \\ \cline{2-3} \cline{5-6} 
                    & S50                  & 0.167        &                       & S50                  & 0.100      \\ \hline
Target Domain       & Other Target Domains & MMD values & Target Domain         & Other Target Domains & MMD values \\ \hline
\multirow{4}{*}{S5} & S3                   & 0.110      & \multirow{4}{*}{S49}  & S3                   & 0.117      \\ \cline{2-3} \cline{5-6} 
                    & S7                   & 0.164      &                       & S7                   & 0.216      \\ \cline{2-3} \cline{5-6} 
                    & S19                  & 0.125      &                       & S19                  & 0.170      \\ \cline{2-3} \cline{5-6} 
                    & S50                  & 0.150      &                       & S50                  & 0.158      \\ \hline
Target Domain       & Other Target Domains & MMD values & Target Domain         & Other Target Domains & MMD values \\ \hline
\multirow{4}{*}{S8} & S3                   & 0.230      & \multirow{4}{*}{NIST} & S3                   & 0.069      \\ \cline{2-3} \cline{5-6} 
                    & S7                   & 0.083      &                       & S7                   & 0.170      \\ \cline{2-3} \cline{5-6} 
                    & S19                  & 0.088      &                       & S19                  & 0.123      \\ \cline{2-3} \cline{5-6} 
                    & S50                  & 0.240      &                       & S50                  & 0.110      \\ \hline
\end{tabular}
\end{table*}
\begin{table*}[t!]
\caption{Calculation of MMD values between source and target domains for MORE data.}
\centering
\label{tab:mmd_more}
\begin{tabular}{|c|c|c|c|c|c|}
\hline
Source Domain & Other Target Domains & MMD values & Target Domain & Other Target Domains & MMD values \\ \hline
\multirow{4}{*}{WT11} & WT2 & 0.079 & \multirow{4}{*}{WT4} & WT2 & 0.077 \\ \cline{2-3} \cline{5-6} 
 & WT3 & 0.084 &  & WT3 & 0.084 \\ \cline{2-3} \cline{5-6} 
 & WT5 & 0.095 &  & WT5 & 0.092 \\ \cline{2-3} \cline{5-6} 
 & WT9 & 0.095 &  & WT9 & 0.093 \\ \hline
Target Domain & Other Target Domains & MMD values & Target Domain & Other Target Domains & MMD values \\ \hline
\multirow{4}{*}{WT6} & WT2 & 0.080 & \multirow{4}{*}{WT7} & WT2 & 0.079 \\ \cline{2-3} \cline{5-6} 
 & WT3 & 0.081 &  & WT3 & 0.093 \\ \cline{2-3} \cline{5-6} 
 & WT5 & 0.095 &  & WT5 & 0.097 \\ \cline{2-3} \cline{5-6} 
 & WT9 & 0.100 &  & WT9 & 0.095 \\ \hline
Target Domain & Other Target Domains & MMD values & Target Domain & Other Target Domains & MMD values \\ \hline
\multirow{4}{*}{WT8} & WT2 & 0.085 & \multirow{4}{*}{WT9} & WT2 & 0.080 \\ \cline{2-3} \cline{5-6} 
 & WT3 & 0.085 &  & WT3 & 0.083 \\ \cline{2-3} \cline{5-6} 
 & WT5 & 0.100 &  & WT5 & 0.098 \\ \cline{2-3} \cline{5-6} 
 & WT9 & 0.099 &  & WT9 & 0.100 \\ \hline
\end{tabular}
\end{table*}
\subsection{Dataset Overview}
This section provides an overview of the training and evaluation datasets used during the pre-training and fine-tuning phases for Energy and MORE data. Table~\ref{tab:dataset} shows the amount of data used for training and testing the source model and fine-tuned target models. Here, we can observe that the source domains S4 and WT11 have sufficient model training data compared to the target domains, even after we add the certain amount of data from source domain to the target domains. 

\subsection{Hyper-parameters used during pre-training}
Table~\ref{tab:param_source} shows the summary of parameters used during pre-training the source model in the selected source domains S4 and WT11 for Energy and MORE data, respectively. The hyper-parameters are obtained through grid search.

\subsection{Ablation Study}\label{ablation}
We also performed additional experiments on Energy data for an ablation study of our design decisions. In Fig~\ref{fig:varyingpercent}, we show how adding a certain amount of data samples from the source domain affects the performance of target domains during source model fine-tuning. As can be seen, for the target domains S5, S49, and NIST, adding 5\% of data from the source domain performs better compared to other percentages of added data, as the MMD values between these target domains with S4 are higher (see Table~\ref{tab:mmd1}). Likewise, for the target domains S8 and S15, adding 20\% of data from the source domain works best compared to other percentages of added data, as the MMD values between these target domains with S4 are lower (see Table~\ref{tab:mmd1}). 

Next, in Fig~\ref{fig:freezing_tech}, we experimented with various techniques for freezing pre-trained layers. We compared our \emph{One-step fine-tuning} approach with two different methods: \textbf{Top layer unfreeze} and \textbf{w/o GU}. For the \textbf{Top Layer Unfreeze approach}, we only unfreeze the top layer- specifically, the decoder layer of the source model. We then fine-tune the model using the GU technique. In the \textbf{w/o GU approach}, we unfreeze all the layers at once at the initial stage of fine-tuning rather than gradually unfreezing the source model layers. Our \emph{One-step fine-tuning} approach, where we gradually unfreeze all the pre-trained layers after adding some percentage of source domain data to the target domain, performs better than the other two approaches. Here, we can see an improvement of up to 40\%. 

\subsection{Distribution Alignment Analysis}
In this section, we delve into a thorough analysis of the alignment between the data distributions of the source and different target domains. Table~\ref{tab:mmd1}, Table~\ref{tab:mmd}, and Table~\ref{tab:mmd_more} present the MMD values, which serve as a crucial metric for quantifying the dissimilarity or similarity between these domains. 

Table~\ref{tab:mmd1} shows the calculation of MMD values between the source and target domains for Energy data. These target domains were selected for source model fine-tuning. Here, the varying MMD values among the target domains, ranging from 0.089 to 0.133, show the discrepancies in the alignment between the source and individual target domains. Notably, the lower MMD values, such as 0.089, indicate a more effective alignment, indicating a closer resemblance in the feature distributions between the source and those specific target domains. This potentially signifies more robust adaptability of the model to those domains. On the other hand, higher MMD values, like 0.133, imply a more significant dissimilarity, highlighting challenges in aligning certain target domains with the source. 

Next, Table~\ref{tab:mmd} and Table ~\ref{tab:mmd_more} shows the calculation of MMD values between the source and various target domains for Energy and MORE data, respectively. Here, the other target domains were not involved during source model fine-tuning. The tables not only provide the perspective of the alignment between the source and individual target domains but also of the disparities among the various target domains themselves. These MMD values help provide insight into examining data shifts between the source and different target domains in the domain adaptation process. This analysis allows us to see how well the domain adaptation strategy works and how the fine-tuned target model works well across different domains.

\end{document}

%% file: sdavsp.tex
\begin{figure}[t!]
\begin{subfigure}[b]{0.20\textwidth}
\centering
% This file was created with tikzplotlib v0.10.1.
\begin{tikzpicture}[scale=0.5]
\pgfplotsset{
every axis legend/.append style={at={(0.5,1.1)}, anchor=center,legend columns = 2},
legend style={/tikz/every even column/.append style={column sep=0.5cm}}}

\definecolor{darkslategray38}{RGB}{38,38,38}
\definecolor{lightgray204}{RGB}{204,204,204}
\definecolor{peru22113282}{RGB}{221,132,82}
\definecolor{steelblue76114176}{RGB}{76,114,176}
\definecolor{royalblue}{RGB}{65,105,225}

\begin{axis}[
tick align=outside,
tick pos=left,
x grid style={lightgray204},
xlabel={\textbf{Portion of Test Data Samples}},
xmajorgrids,
xmin=-9.95, xmax=208.95,
xtick style={color=darkslategray38},
y grid style={lightgray204},
ylabel={\textbf{Indoor Temperature}},
ymajorgrids,
ymin=19.4732629776001, ymax=25.5616033554077,
ytick style={color=darkslategray38}
]
\addlegendimage{area legend,pattern=north west lines, color=peru22113282,draw=white}
\addlegendimage{area legend,pattern=north west lines, color=royalblue,draw=white}
\addlegendentry{Actual (S4)}
\addlegendentry{Predicted (S4)}

\addplot [very thick, peru22113282]
table {%
0 22.416259765625
1 22.5324058532715
2 21.9602470397949
3 22.6665420532227
4 21.9826545715332
5 24.253231048584
6 21.8394317626953
7 22.6443099975586
8 22.6966724395752
9 22.1502418518066
10 22.0373859405518
11 22.2731513977051
12 22.9846458435059
13 22.1749019622803
14 22.1433048248291
15 22.513801574707
16 22.4837989807129
17 22.4905414581299
18 21.8076801300049
19 22.5332260131836
20 22.6950912475586
21 21.6562538146973
22 22.6662883758545
23 23.5961227416992
24 22.2703895568848
25 22.6434726715088
26 22.3575077056885
27 22.7060985565186
28 21.7655029296875
29 23.7383117675781
30 22.7000026702881
31 22.0357799530029
32 22.1518135070801
33 22.150562286377
34 21.7948265075684
35 21.7894535064697
36 22.1523170471191
37 22.3897533416748
38 21.8793601989746
39 21.7618503570557
40 22.2728309631348
41 21.971996307373
42 22.1234283447266
43 22.1451263427734
44 22.6859397888184
45 22.1649036407471
46 21.9650421142578
47 22.1368942260742
48 23.034517288208
49 22.8792724609375
50 22.0832672119141
51 22.170431137085
52 21.6767272949219
53 22.2948455810547
54 21.7768535614014
55 22.5145626068115
56 22.561544418335
57 22.0847434997559
58 22.6440830230713
59 21.4642696380615
60 21.7924098968506
61 22.5854396820068
62 22.1245574951172
63 22.0872440338135
64 23.6247253417969
65 19.7500057220459
66 22.646598815918
67 21.5804004669189
68 22.4008769989014
69 22.6794414520264
70 22.1375198364258
71 22.0438594818115
72 22.670166015625
73 21.8157024383545
74 22.6398735046387
75 22.7227077484131
76 21.3477249145508
77 22.381872177124
78 25.239631652832
79 22.7283725738525
80 22.2430458068848
81 22.2532157897949
82 22.3136692047119
83 22.277322769165
84 21.8974876403809
85 22.3860626220703
86 22.1414928436279
87 22.676155090332
88 23.4402008056641
89 22.2203884124756
90 22.158763885498
91 22.33620262146
92 22.6499900817871
93 22.5912380218506
94 22.6392211914062
95 22.6546211242676
96 22.408109664917
97 21.9437713623047
98 22.6407508850098
99 22.7449188232422
100 21.999626159668
101 22.6891422271729
102 22.0347747802734
103 22.6452407836914
104 22.1576976776123
105 22.376932144165
106 21.7994327545166
107 22.0231971740723
108 22.1736545562744
109 21.9833564758301
110 22.5353355407715
111 22.1092109680176
112 22.1518135070801
113 22.1516513824463
114 22.3418827056885
115 22.7725410461426
116 22.1770668029785
117 22.0113964080811
118 22.9646739959717
119 21.8980407714844
120 22.4587326049805
121 22.6467800140381
122 21.937313079834
123 22.0243988037109
124 22.5225734710693
125 21.9780254364014
126 22.8060646057129
127 21.7877349853516
128 23.2043495178223
129 22.1754493713379
130 22.415641784668
131 22.5914115905762
132 22.300666809082
133 22.8882389068604
134 23.0624217987061
135 22.1517906188965
136 23.034517288208
137 22.5787906646729
138 22.6572914123535
139 22.1302070617676
140 22.621452331543
141 22.0798664093018
142 22.143404006958
143 22.644718170166
144 22.637092590332
145 22.1482391357422
146 22.490650177002
147 22.7166423797607
148 22.7000789642334
149 22.2954769134521
150 22.8227691650391
151 21.8019466400146
152 22.1941299438477
153 22.7836494445801
154 22.6569328308105
155 22.696216583252
156 21.5817375183105
157 22.1657371520996
158 21.7917175292969
159 23.028392791748
160 22.9434223175049
161 21.8705024719238
162 21.9872741699219
163 22.0087947845459
164 22.8407669067383
165 22.5035552978516
166 22.5729789733887
167 22.8499412536621
168 21.9100532531738
169 21.8829898834229
170 22.2020320892334
171 22.5015048980713
172 21.5434646606445
173 22.023983001709
174 22.6697883605957
175 22.41623878479
176 22.9485111236572
177 22.4963092803955
178 22.7724361419678
179 22.6744937896729
180 22.1932411193848
181 22.4601287841797
182 21.5864734649658
183 22.636344909668
184 21.9878711700439
185 22.3887882232666
186 22.8273811340332
187 22.5874996185303
188 22.307409286499
189 22.700719833374
190 21.8130321502686
191 22.8119621276855
192 22.670166015625
193 22.2336463928223
194 22.7119522094727
195 22.16676902771
196 21.695592880249
197 22.6812629699707
198 22.2415504455566
199 22.2424449920654
};

\addplot [very thick, royalblue]
table {%
0 22.4844455718994
1 22.4490280151367
2 21.984582901001
3 22.6665420532227
4 22.1156940460205
5 24.4444446563721
6 21.6666660308838
7 22.7777786254883
8 22.7777786254883
9 22.0401382446289
10 22.2222232818604
11 22.2222232818604
12 22.9398612976074
13 22.2222232818604
14 22.1433048248291
15 22.6002788543701
16 22.4119434356689
17 22.3240280151367
18 21.6666660308838
19 22.7777786254883
20 22.7777786254883
21 21.3055572509766
22 22.8873615264893
23 23.3549995422363
24 22.4738883972168
25 22.320972442627
26 22.3502788543701
27 22.7600002288818
28 21.5293045043945
29 24.2808322906494
30 22.7777786254883
31 22.2222232818604
32 21.9398612976074
33 22.7777786254883
34 21.6666660308838
35 21.7762489318848
36 22.2222232818604
37 22.3009738922119
38 22.2222232818604
39 21.6666660308838
40 22.2728309631348
41 21.971996307373
42 22.1234283447266
43 22.1451263427734
44 22.6859397888184
45 22.1649036407471
46 21.9650421142578
47 22.1368942260742
48 23.034517288208
49 22.8792724609375
50 22.0832672119141
51 22.170431137085
52 21.6767272949219
53 22.2948455810547
54 21.7768535614014
55 22.5145626068115
56 22.561544418335
57 22.0847434997559
58 22.6440830230713
59 21.4642696380615
60 21.7924098968506
61 22.5854396820068
62 22.1245574951172
63 22.0872440338135
64 23.6247253417969
65 19.7500057220459
66 22.646598815918
67 21.5804004669189
68 22.4008769989014
69 22.6794414520264
70 22.1375198364258
71 22.0438594818115
72 22.670166015625
73 21.8157024383545
74 22.6398735046387
75 22.7227077484131
76 21.3477249145508
77 22.381872177124
78 25.239631652832
79 22.7283725738525
80 22.2430458068848
81 22.2532157897949
82 22.3136692047119
83 22.277322769165
84 21.8974876403809
85 22.3860626220703
86 22.1414928436279
87 22.676155090332
88 23.4402008056641
89 22.2203884124756
90 22.158763885498
91 22.33620262146
92 22.6499900817871
93 22.5912380218506
94 22.6392211914062
95 22.6546211242676
96 22.408109664917
97 21.9437713623047
98 22.6407508850098
99 22.7449188232422
100 21.999626159668
101 22.413610458374
102 22.2222232818604
103 23.0433330535889
104 22.2222232818604
105 22.4659729003906
106 21.6666660308838
107 22.2129173278809
108 22.7037506103516
109 22.211389541626
110 22.3394451141357
111 22.5633335113525
112 22.0586128234863
113 22.461389541626
114 22.534029006958
115 22.864444732666
116 22.2222232818604
117 21.9645843505859
118 23.0216674804688
119 21.924165725708
120 22.7777786254883
121 22.804027557373
122 22.0216655731201
123 22.0431938171387
124 22.5215282440186
125 22.0793056488037
126 23.1990280151367
127 21.6666660308838
128 23.8888893127441
129 22.2222232818604
130 22.415641784668
131 22.5914115905762
132 22.300666809082
133 22.8882389068604
134 23.0624217987061
135 22.1517906188965
136 23.034517288208
137 22.5787906646729
138 22.6572914123535
139 22.1302070617676
140 22.621452331543
141 22.0798664093018
142 22.143404006958
143 22.644718170166
144 22.637092590332
145 22.1482391357422
146 22.490650177002
147 22.7166423797607
148 22.7000789642334
149 22.2954769134521
150 22.8227691650391
151 21.6805553436279
152 22.3641662597656
153 23.125
154 22.6681957244873
155 23.0247211456299
156 20.8780555725098
157 22.3961124420166
158 21.6666660308838
159 23.3333320617676
160 23.6187477111816
161 21.6666660308838
162 22.1759738922119
163 22.1295852661133
164 23.3333320617676
165 22.3348617553711
166 22.6574993133545
167 22.9645843505859
168 21.6666660308838
169 22.2222232818604
170 22.4027767181396
171 22.5015048980713
172 21.5434646606445
173 22.023983001709
174 22.6697883605957
175 22.41623878479
176 22.9485111236572
177 22.4963092803955
178 22.7724361419678
179 22.6744937896729
180 22.1932411193848
181 22.4601287841797
182 21.5864734649658
183 22.636344909668
184 21.9878711700439
185 22.3887882232666
186 22.8273811340332
187 22.5874996185303
188 22.307409286499
189 22.700719833374
190 21.8130321502686
191 22.8119621276855
192 22.670166015625
193 22.2336463928223
194 22.7119522094727
195 22.16676902771
196 21.695592880249
197 22.6812629699707
198 22.2415504455566
199 22.2424449920654
};
\end{axis}

\end{tikzpicture}
\caption{}
\label{fig:sd_wt}
\end{subfigure}\hspace{2em}
\begin{subfigure}[b]{0.20\textwidth}
\centering
% This file was created with tikzplotlib v0.10.1.
\begin{tikzpicture}[scale=0.5]
\pgfplotsset{
every axis legend/.append style={at={(0.5,1.1)}, anchor=center,legend columns = 2},
legend style={/tikz/every even column/.append style={column sep=0.5cm}}}

\definecolor{darkgray176}{RGB}{176,176,176}
\definecolor{darkorange25512714}{RGB}{255,127,14}
\definecolor{steelblue31119180}{RGB}{31,119,180}
\definecolor{royalblue}{RGB}{65,105,225}

\begin{axis}[
tick align=outside,
tick pos=left,
x grid style={darkgray176},
xmajorgrids,
xlabel={\textbf{Portion of Test Data Samples}},
xmin=-9.95, xmax=208.95,
xtick style={color=black},
y grid style={darkgray176},
ymajorgrids,
ylabel={\textbf{Wind Power}},
ymin=-7.28323149681091, ymax=43.2870166301727,
ytick style={color=black}
]
\addlegendimage{area legend,pattern=north west lines, color=darkorange25512714,draw=white}
\addlegendimage{area legend,pattern=north west lines, color=royalblue,draw=white}
\addlegendentry{Actual (WT11)}
\addlegendentry{Predicted (WT11)}

\addplot [very thick, darkorange25512714]
table {%
0 20.2273921966553
1 18.7596912384033
2 17.2305488586426
3 11.5552568435669
4 11.5960941314697
5 11.8880338668823
6 12.9844026565552
7 16.639684677124
8 18.019157409668
9 17.2868461608887
10 16.771146774292
11 18.1654796600342
12 15.009838104248
13 12.927544593811
14 7.90215253829956
15 9.82115840911865
16 13.8464622497559
17 12.9533452987671
18 10.4712343215942
19 8.45495128631592
20 9.11612606048584
21 5.02820634841919
22 3.29399609565735
23 0.830555438995361
24 0.729791700839996
25 -0.472687602043152
26 -0.822189271450043
27 -1.24041318893433
28 -0.835636496543884
29 0.0167042016983032
30 0.610412418842316
31 4.85626697540283
32 3.19941234588623
33 2.90129852294922
34 5.54083204269409
35 3.81906032562256
36 1.58984994888306
37 1.04538547992706
38 0.405372053384781
39 -2.01054120063782
40 -1.65707981586456
41 -4.27801084518433
42 -4.10262775421143
43 -3.72868037223816
44 -2.40437483787537
45 -0.507526457309723
46 -2.02924823760986
47 -2.86709117889404
48 -1.67843008041382
49 -0.102887541055679
50 0.438314944505692
51 1.42370975017548
52 3.47165250778198
53 9.06697368621826
54 14.4078073501587
55 22.8269348144531
56 21.1463356018066
57 20.9305591583252
58 17.4932651519775
59 18.5946712493896
60 13.0266408920288
61 12.005108833313
62 10.9496726989746
63 11.0831890106201
64 12.9024419784546
65 10.0254402160645
66 4.49796009063721
67 4.15142107009888
68 3.48772501945496
69 0.630420088768005
70 -2.52652549743652
71 -3.49410676956177
72 -3.40040516853333
73 -2.73010277748108
74 -2.24842834472656
75 -1.76692891120911
76 -3.70035147666931
77 -2.88640141487122
78 -0.787179052829742
79 6.60622978210449
80 8.66629028320312
81 13.2216873168945
82 13.9471521377563
83 12.9654731750488
84 9.79572296142578
85 13.7818183898926
86 17.1704692840576
87 16.9034805297852
88 11.3068284988403
89 10.1733703613281
90 4.49597835540771
91 9.90708446502686
92 12.4315156936646
93 16.4261627197266
94 15.7583866119385
95 18.9888687133789
96 11.3707084655762
97 8.01471996307373
98 10.64084815979
99 14.3047780990601
100 13.6735553741455
101 16.1267242431641
102 15.599347114563
103 20.153039932251
104 20.0033721923828
105 15.1187868118286
106 16.6999053955078
107 11.4905242919922
108 13.3732137680054
109 11.6726675033569
110 12.9613342285156
111 13.6820955276489
112 17.7611293792725
113 16.0356502532959
114 13.566478729248
115 13.5843391418457
116 18.1198024749756
117 13.595799446106
118 15.8601322174072
119 26.0880889892578
120 29.4823551177979
121 19.6246147155762
122 31.3764877319336
123 33.3515319824219
124 33.941276550293
125 33.748420715332
126 30.6649532318115
127 29.6882457733154
128 26.8704643249512
129 26.5770721435547
130 28.5650978088379
131 39.2224578857422
132 36.4676895141602
133 27.3462333679199
134 20.1944217681885
135 15.4059591293335
136 15.4582834243774
137 26.7037563323975
138 14.8881921768188
139 19.2615184783936
140 25.7218990325928
141 17.8839244842529
142 17.1555595397949
143 14.5915994644165
144 16.6911010742188
145 16.7031364440918
146 16.0522270202637
147 15.2340173721313
148 15.3266353607178
149 12.1093215942383
150 8.65837097167969
151 3.8895947933197
152 7.00730133056641
153 5.5544056892395
154 6.61748361587524
155 6.98627710342407
156 14.3809976577759
157 15.7718124389648
158 12.4393453598022
159 9.79874420166016
160 9.36560916900635
161 13.8707761764526
162 9.49519443511963
163 14.2499742507935
164 24.2053203582764
165 20.6148872375488
166 25.2178230285645
167 27.4420528411865
168 26.2552642822266
169 30.9347190856934
170 30.2553405761719
171 32.6076354980469
172 26.431604385376
173 17.9423999786377
174 18.6089973449707
175 17.6064529418945
176 18.7825241088867
177 18.9970226287842
178 21.3019275665283
179 17.6668663024902
180 23.3694095611572
181 27.276554107666
182 39.4206352233887
183 38.3276748657227
184 40.4045600891113
185 40.9883689880371
186 40.0336227416992
187 36.5759658813477
188 36.6345443725586
189 38.8624877929688
190 38.2014045715332
191 39.0903625488281
192 38.8262901306152
193 40.4600791931152
194 39.9671173095703
195 33.9297485351562
196 40.5730285644531
197 40.3364906311035
198 40.7061958312988
199 40.645450592041
};

\addplot [very thick, royalblue]
table {%
0 20.2273921966553
1 18.7596912384033
2 17.2305488586426
3 11.5552568435669
4 11.5960941314697
5 11.8880338668823
6 12.9844026565552
7 16.639684677124
8 18.019157409668
9 17.2868461608887
10 16.771146774292
11 18.1654796600342
12 15.009838104248
13 12.927544593811
14 7.90215253829956
15 9.82115840911865
16 13.8464622497559
17 12.9533452987671
18 10.4712343215942
19 8.45495128631592
20 9.11612606048584
21 4.16767358779907
22 2.19222712516785
23 -0.369866847991943
24 -1.00016796588898
25 -1.86604928970337
26 -2.52103185653687
27 -2.05502915382385
28 -0.679491639137268
29 -0.214225739240646
30 3.48082590103149
31 1.5877126455307
32 3.82931542396545
33 5.13025522232056
34 3.01703381538391
35 2.63612675666809
36 1.92607164382935
37 1.86768269538879
38 -1.44369959831238
39 -2.34199833869934
40 -4.31928682327271
41 -4.80610322952271
42 -4.8987512588501
43 -3.2227988243103
44 -0.31221541762352
45 -0.507526457309723
46 -2.02924823760986
47 -2.86709117889404
48 -1.67843008041382
49 -0.102887541055679
50 0.438314944505692
51 1.42370975017548
52 3.47165250778198
53 9.06697368621826
54 14.4078073501587
55 22.8269348144531
56 21.1463356018066
57 20.9305591583252
58 17.4932651519775
59 18.5946712493896
60 13.0266408920288
61 12.005108833313
62 10.9496726989746
63 11.0831890106201
64 12.9024419784546
65 10.0254402160645
66 4.49796009063721
67 4.15142107009888
68 3.48772501945496
69 0.630420088768005
70 -2.52652549743652
71 -3.49410676956177
72 -3.40040516853333
73 -2.73010277748108
74 -2.24842834472656
75 -1.76692891120911
76 -3.70035147666931
77 -2.88640141487122
78 -0.787179052829742
79 6.60622978210449
80 8.66629028320312
81 13.2216873168945
82 13.9471521377563
83 12.9654731750488
84 9.79572296142578
85 13.7818183898926
86 17.1704692840576
87 16.9034805297852
88 11.3068284988403
89 10.1733703613281
90 4.49597835540771
91 4.09049797058105
92 10.9129257202148
93 12.51096534729
94 18.1549396514893
95 7.78223514556885
96 3.65764904022217
97 9.40311431884766
98 14.195574760437
99 13.031120300293
100 13.6735553741455
101 16.1267242431641
102 15.599347114563
103 20.153039932251
104 20.0033721923828
105 15.1187868118286
106 16.6999053955078
107 11.4905242919922
108 13.3732137680054
109 11.6726675033569
110 12.9613342285156
111 13.6820955276489
112 17.7611293792725
113 16.0356502532959
114 13.566478729248
115 13.5843391418457
116 18.1198024749756
117 13.595799446106
118 15.8601322174072
119 26.0880889892578
120 29.4823551177979
121 19.6246147155762
122 31.3764877319336
123 33.3515319824219
124 33.941276550293
125 33.748420715332
126 30.6649532318115
127 29.6882457733154
128 26.8704643249512
129 26.5770721435547
130 28.5650978088379
131 39.2224578857422
132 36.4676895141602
133 27.3462333679199
134 20.1944217681885
135 15.4059591293335
136 15.4582834243774
137 26.7037563323975
138 14.8881921768188
139 19.2615184783936
140 25.7218990325928
141 17.8839244842529
142 17.1555595397949
143 14.5915994644165
144 16.6911010742188
145 16.7031364440918
146 16.0522270202637
147 15.2340173721313
148 15.3266353607178
149 12.1093215942383
150 8.65837097167969
151 3.8895947933197
152 7.00730133056641
153 5.5544056892395
154 6.61748361587524
155 6.98627710342407
156 14.3809976577759
157 15.7718124389648
158 12.4393453598022
159 9.79874420166016
160 9.36560916900635
161 13.8707761764526
162 9.49519443511963
163 14.2499742507935
164 24.2053203582764
165 20.6148872375488
166 25.2178230285645
167 27.4420528411865
168 26.2552642822266
169 30.9347190856934
170 30.2553405761719
171 26.2534408569336
172 19.4035701751709
173 19.012716293335
174 13.8686361312866
175 14.6035575866699
176 16.2885341644287
177 17.5701351165771
178 15.2590923309326
179 23.1444854736328
180 24.4562931060791
181 33.3593406677246
182 29.1776542663574
183 34.0058479309082
184 38.5513076782227
185 37.7920761108398
186 32.5929222106934
187 31.5725059509277
188 35.7512397766113
189 34.619571685791
190 38.2014045715332
191 39.0903625488281
192 38.8262901306152
193 40.4600791931152
194 39.9671173095703
195 33.9297485351562
196 40.5730285644531
197 40.3364906311035
198 40.7061958312988
199 40.645450592041
};
\end{axis}

\end{tikzpicture}
\caption{}
\label{fig:sd_hp}
\end{subfigure}
\caption{Performance of Source Model: Actual v.s. Predicted values of Source Domains: (a) S4, and (b) WT11.}
    \label{fig:avsp_sd}
\end{figure}
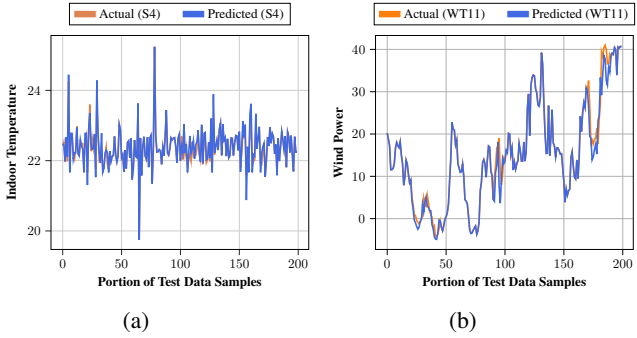

%% file: baselines.tex
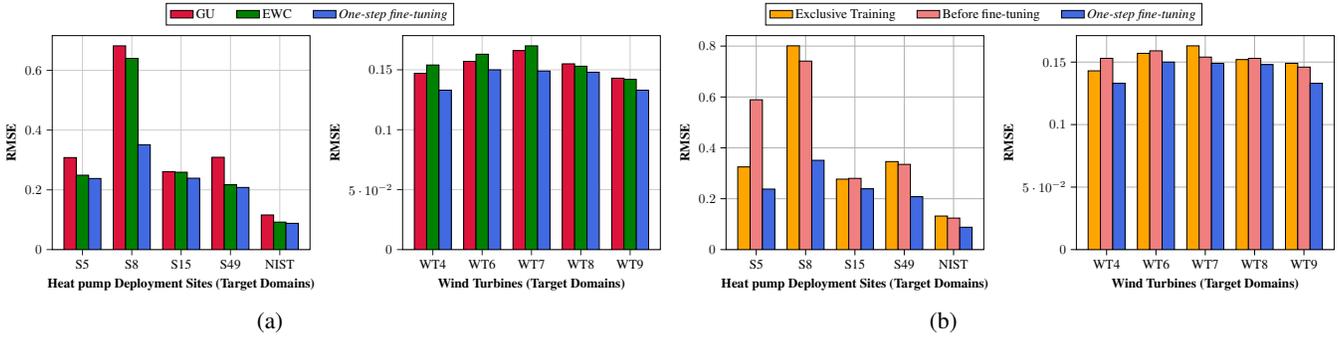
\begin{figure*}[t!]
\begin{subfigure}[b]{0.4\textwidth}
    \centering
% This file was created with tikzplotlib v0.10.1.
\begin{tikzpicture}[scale=0.5]
\pgfplotsset{
every axis legend/.append style={at={(1,1.1)}, anchor=center,legend columns = 3},
legend style={/tikz/every even column/.append style={column sep=0.5cm}}}

\definecolor{crimson}{RGB}{220,20,60}
\definecolor{green}{RGB}{0,128,0}
\definecolor{darkslategray38}{RGB}{38,38,38}
\definecolor{lightgray204}{RGB}{204,204,204}
\definecolor{royalblue}{RGB}{65,105,225}

\begin{groupplot}[group style={group size=2 by 1, vertical sep =20pt, horizontal sep=70pt}]
\nextgroupplot[
tick align=outside,
tick pos=left,
x grid style={lightgray204},
xlabel={\textbf{Heat pump Deployment Sites (Target Domains)}},
xmajorgrids,
xmin=-0.3625, xmax=4.8625,
xtick style={color=darkslategray38},
xtick={0.25,1.25,2.25,3.25,4.25},
xticklabels={S5,S8,S15,S49,NIST},
y grid style={lightgray204},
ylabel={\textbf{RMSE}},
ymajorgrids,
ymin=0, ymax=0.7161,
ytick style={color=darkslategray38}
]
\addlegendimage{area legend,pattern=north west lines, color=crimson,draw=black}
\addlegendimage{area legend,pattern=north west lines, color=green,draw=black}
\addlegendimage{area legend,pattern=north west lines, color=royalblue,draw=black}
\addlegendentry{GU}
\addlegendentry{EWC}
\addlegendentry{\emph{One-step fine-tuning}}

\draw[draw=black,fill=crimson] (axis cs:-0.125,0) rectangle (axis cs:0.125,0.308);
\draw[draw=black,fill=crimson] (axis cs:0.875,0) rectangle (axis cs:1.125,0.682);
\draw[draw=black,fill=crimson] (axis cs:1.875,0) rectangle (axis cs:2.125,0.261);
\draw[draw=black,fill=crimson] (axis cs:2.875,0) rectangle (axis cs:3.125,0.309);
\draw[draw=black,fill=crimson] (axis cs:3.875,0) rectangle (axis cs:4.125,0.116);
\draw[draw=black,fill=green] (axis cs:0.125,0) rectangle (axis cs:0.375,0.249);
\draw[draw=black,fill=green] (axis cs:1.125,0) rectangle (axis cs:1.375,0.64);
\draw[draw=black,fill=green] (axis cs:2.125,0) rectangle (axis cs:2.375,0.259);
\draw[draw=black,fill=green] (axis cs:3.125,0) rectangle (axis cs:3.375,0.217);
\draw[draw=black,fill=green] (axis cs:4.125,0) rectangle (axis cs:4.375,0.092);
\draw[draw=black,fill=royalblue] (axis cs:0.375,0) rectangle (axis cs:0.625,0.238);
\draw[draw=black,fill=royalblue] (axis cs:1.375,0) rectangle (axis cs:1.625,0.351);
\draw[draw=black,fill=royalblue] (axis cs:2.375,0) rectangle (axis cs:2.625,0.239);
\draw[draw=black,fill=royalblue] (axis cs:3.375,0) rectangle (axis cs:3.625,0.208);
\draw[draw=black,fill=royalblue] (axis cs:4.375,0) rectangle (axis cs:4.625,0.088);

\nextgroupplot[
tick align=outside,
tick pos=left,
x grid style={lightgray204},
xlabel={\textbf{Wind Turbines (Target Domains)}},
xmajorgrids,
xmin=-0.3625, xmax=4.8625,
xtick style={color=darkslategray38},
xtick={0.25,1.25,2.25,3.25,4.25},
xticklabels={WT4,WT6,WT7,WT8,WT9},
y grid style={lightgray204},
ylabel={\textbf{RMSE}},
ymajorgrids,
ymin=0, ymax=0.1785,
ytick style={color=darkslategray38}
]
\draw[draw=black,fill=crimson] (axis cs:-0.125,0) rectangle (axis cs:0.125,0.147);
\draw[draw=black,fill=crimson] (axis cs:0.875,0) rectangle (axis cs:1.125,0.157);
\draw[draw=black,fill=crimson] (axis cs:1.875,0) rectangle (axis cs:2.125,0.166);
\draw[draw=black,fill=crimson] (axis cs:2.875,0) rectangle (axis cs:3.125,0.155);
\draw[draw=black,fill=crimson] (axis cs:3.875,0) rectangle (axis cs:4.125,0.143);
\draw[draw=black,fill=green] (axis cs:0.125,0) rectangle (axis cs:0.375,0.154);
\draw[draw=black,fill=green] (axis cs:1.125,0) rectangle (axis cs:1.375,0.163);
\draw[draw=black,fill=green] (axis cs:2.125,0) rectangle (axis cs:2.375,0.17);
\draw[draw=black,fill=green] (axis cs:3.125,0) rectangle (axis cs:3.375,0.153);
\draw[draw=black,fill=green] (axis cs:4.125,0) rectangle (axis cs:4.375,0.142);
\draw[draw=black,fill=royalblue] (axis cs:0.375,0) rectangle (axis cs:0.625,0.133);
\draw[draw=black,fill=royalblue] (axis cs:1.375,0) rectangle (axis cs:1.625,0.15);
\draw[draw=black,fill=royalblue] (axis cs:2.375,0) rectangle (axis cs:2.625,0.149);
\draw[draw=black,fill=royalblue] (axis cs:3.375,0) rectangle (axis cs:3.625,0.148);
\draw[draw=black,fill=royalblue] (axis cs:4.375,0) rectangle (axis cs:4.625,0.133);
\end{groupplot}

\end{tikzpicture}

\caption{}
 \label{fig:ftm}
\end{subfigure}\hspace{5em}
\begin{subfigure}[b]{0.4\textwidth}
    \centering
% This file was created with tikzplotlib v0.10.1.
\begin{tikzpicture}[scale=0.5]
\pgfplotsset{
every axis legend/.append style={at={(1,1.1)}, anchor=center,legend columns = 3},
legend style={/tikz/every even column/.append style={column sep=0.5cm}}}

\definecolor{darkgray176}{RGB}{176,176,176}
\definecolor{lightcoral}{RGB}{240,128,128}
\definecolor{orange}{RGB}{255,165,0}
\definecolor{royalblue}{RGB}{65,105,225}

\begin{groupplot}[group style={group size=2 by 1, vertical sep =20pt, horizontal sep=70pt}]
\nextgroupplot[
tick align=outside,
tick pos=left,
x grid style={darkgray176},
xlabel={\textbf{Heat pump Deployment Sites (Target Domains)}},
xmin=-0.3625, xmax=4.8625,
xmajorgrids,
xtick style={color=black},
xtick={0.25,1.25,2.25,3.25,4.25},
xticklabels={S5,S8,S15,S49,NIST},
y grid style={darkgray176},
ylabel={\textbf{RMSE}},
ymajorgrids,
ymin=0, ymax=0.84105,
ytick style={color=black}
]
\addlegendimage{area legend,pattern=north west lines, color=orange,draw=black}
\addlegendimage{area legend,pattern=north west lines, color=lightcoral,draw=black}
\addlegendimage{area legend,pattern=north west lines, color=royalblue,draw=black}
\addlegendentry{Exclusive Training}
\addlegendentry{Before fine-tuning}
\addlegendentry{\emph{One-step fine-tuning}}

\draw[draw=black,fill=orange] (axis cs:-0.125,0) rectangle (axis cs:0.125,0.325);
\draw[draw=black,fill=orange] (axis cs:0.875,0) rectangle (axis cs:1.125,0.801);
\draw[draw=black,fill=orange] (axis cs:1.875,0) rectangle (axis cs:2.125,0.277);
\draw[draw=black,fill=orange] (axis cs:2.875,0) rectangle (axis cs:3.125,0.346);
\draw[draw=black,fill=orange] (axis cs:3.875,0) rectangle (axis cs:4.125,0.132);
\draw[draw=black,fill=lightcoral] (axis cs:0.125,0) rectangle (axis cs:0.375,0.589);
\draw[draw=black,fill=lightcoral] (axis cs:1.125,0) rectangle (axis cs:1.375,0.741);
\draw[draw=black,fill=lightcoral] (axis cs:2.125,0) rectangle (axis cs:2.375,0.28);
\draw[draw=black,fill=lightcoral] (axis cs:3.125,0) rectangle (axis cs:3.375,0.335);
\draw[draw=black,fill=lightcoral] (axis cs:4.125,0) rectangle (axis cs:4.375,0.124);
\draw[draw=black,fill=royalblue] (axis cs:0.375,0) rectangle (axis cs:0.625,0.238);
\draw[draw=black,fill=royalblue] (axis cs:1.375,0) rectangle (axis cs:1.625,0.351);
\draw[draw=black,fill=royalblue] (axis cs:2.375,0) rectangle (axis cs:2.625,0.239);
\draw[draw=black,fill=royalblue] (axis cs:3.375,0) rectangle (axis cs:3.625,0.208);
\draw[draw=black,fill=royalblue] (axis cs:4.375,0) rectangle (axis cs:4.625,0.088);

\nextgroupplot[
tick align=outside,
tick pos=left,
x grid style={darkgray176},
xlabel={\textbf{Wind Turbines (Target Domains)}},
xmajorgrids,
xmin=-0.3625, xmax=4.8625,
xtick style={color=black},
xtick={0.25,1.25,2.25,3.25,4.25},
xticklabels={WT4,WT6,WT7,WT8,WT9},
y grid style={darkgray176},
ylabel={\textbf{RMSE}},
ymajorgrids,
ymin=0, ymax=0.17115,
ytick style={color=black}
]
\draw[draw=black,fill=orange] (axis cs:-0.125,0) rectangle (axis cs:0.125,0.143);
\draw[draw=black,fill=orange] (axis cs:0.875,0) rectangle (axis cs:1.125,0.157);
\draw[draw=black,fill=orange] (axis cs:1.875,0) rectangle (axis cs:2.125,0.163);
\draw[draw=black,fill=orange] (axis cs:2.875,0) rectangle (axis cs:3.125,0.152);
\draw[draw=black,fill=orange] (axis cs:3.875,0) rectangle (axis cs:4.125,0.149);
\draw[draw=black,fill=lightcoral] (axis cs:0.125,0) rectangle (axis cs:0.375,0.153);
\draw[draw=black,fill=lightcoral] (axis cs:1.125,0) rectangle (axis cs:1.375,0.159);
\draw[draw=black,fill=lightcoral] (axis cs:2.125,0) rectangle (axis cs:2.375,0.154);
\draw[draw=black,fill=lightcoral] (axis cs:3.125,0) rectangle (axis cs:3.375,0.153);
\draw[draw=black,fill=lightcoral] (axis cs:4.125,0) rectangle (axis cs:4.375,0.146);
\draw[draw=black,fill=royalblue] (axis cs:0.375,0) rectangle (axis cs:0.625,0.133);
\draw[draw=black,fill=royalblue] (axis cs:1.375,0) rectangle (axis cs:1.625,0.15);
\draw[draw=black,fill=royalblue] (axis cs:2.375,0) rectangle (axis cs:2.625,0.149);
\draw[draw=black,fill=royalblue] (axis cs:3.375,0) rectangle (axis cs:3.625,0.148);
\draw[draw=black,fill=royalblue] (axis cs:4.375,0) rectangle (axis cs:4.625,0.133);
\end{groupplot}

\end{tikzpicture}

\caption{}
\label{fig:eba}
\end{subfigure}
    \caption{Prediction error of \emph{One-step fine-tuning} compared to (a) fine-tuning baselines, (b) model training baselines.}
    \label{fig:enter-label}
    
\end{figure*}

%% file: eavsp.tex
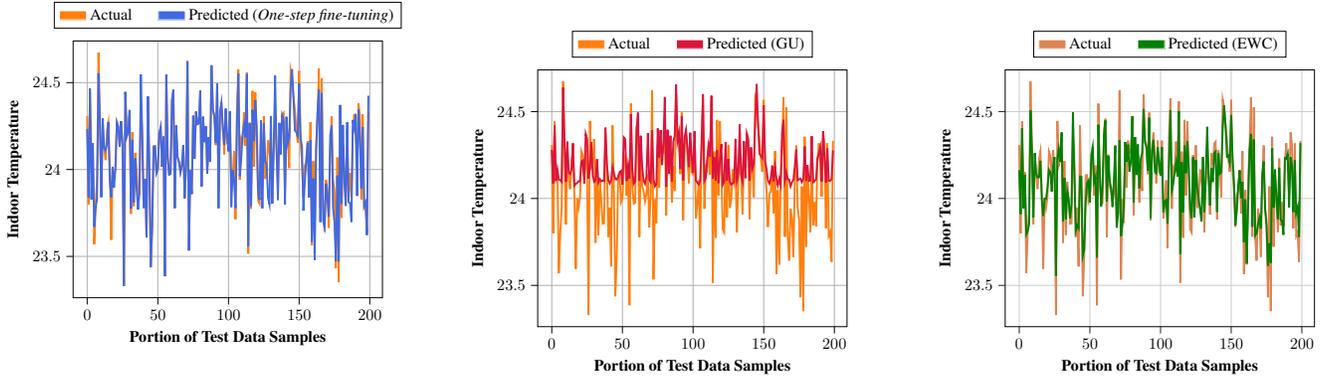
\begin{figure*}[t!]
\centering
\begin{subfigure}[b]{0.30\textwidth}
\centering
% This file was created with tikzplotlib v0.10.1.
\begin{tikzpicture}[scale=0.6]
\pgfplotsset{
every axis legend/.append style={at={(0.5,1.1)}, anchor=center,legend columns = 2},
legend style={/tikz/every even column/.append style={column sep=0.5cm}}}

\definecolor{darkgray176}{RGB}{176,176,176}
\definecolor{darkorange25512714}{RGB}{255,127,14}
\definecolor{steelblue31119180}{RGB}{31,119,180}
\definecolor{royalblue}{RGB}{65,105,225}

\begin{axis}[
tick align=outside,
tick pos=left,
x grid style={darkgray176},
xlabel={\textbf{Portion of Test Data Samples}},
xmajorgrids,
xmin=-9.95, xmax=208.95,
xtick style={color=black},
y grid style={darkgray176},
ylabel={\textbf{Indoor Temperature}},
ymajorgrids,
ymin=23.2621487617493, ymax=24.7397261619568,
ytick style={color=black}
]
\addlegendimage{area legend,pattern=north west lines, color=darkorange25512714,draw=white}
\addlegendimage{area legend,pattern=north west lines, color=royalblue,draw=white}
\addlegendentry{Actual}
\addlegendentry{Predicted (\emph{One-step fine-tuning})}

\addplot [very thick, darkorange25512714]
table {%
0 24.307580947876
1 23.7987747192383
2 24.4434185028076
3 24.0948219299316
4 24.1505222320557
5 23.5697498321533
6 23.8004264831543
7 23.9428977966309
8 24.6725635528564
9 24.2956047058105
10 23.8504047393799
11 24.2559108734131
12 24.124116897583
13 24.0542602539062
14 24.224328994751
15 24.2713832855225
16 24.0202140808105
17 23.5954093933105
18 24.000301361084
19 23.9467964172363
20 24.009822845459
21 24.2718830108643
22 24.2460136413574
23 24.1309700012207
24 24.2785224914551
25 24.1430473327637
26 23.3293113708496
27 24.4460391998291
28 24.1900215148926
29 24.2146835327148
30 24.3423080444336
31 23.7450923919678
32 24.2156562805176
33 23.7900829315186
34 24.0960960388184
35 23.9963130950928
36 23.8544940948486
37 23.8570175170898
38 24.3753395080566
39 24.1999740600586
40 23.7790794372559
41 23.9464340209961
42 23.6088256835938
43 24.420352935791
44 24.0284080505371
45 23.437162399292
46 23.6685485839844
47 24.1327209472656
48 24.1343650817871
49 23.9788265228271
50 24.0013694763184
51 24.2073211669922
52 24.2439804077148
53 24.0156345367432
54 24.1892051696777
55 23.3855266571045
56 24.5460529327393
57 24.0738945007324
58 23.9704608917236
59 23.9746627807617
60 24.3997783660889
61 24.4584407806396
62 23.7780456542969
63 24.2528457641602
64 24.0702972412109
65 24.0424880981445
66 23.9833316802979
67 24.0257053375244
68 24.1411151885986
69 23.8339996337891
70 24.0256366729736
71 24.6232414245605
72 23.5338439941406
73 23.9950771331787
74 23.8591575622559
75 24.3092708587646
76 24.0621585845947
77 24.3329238891602
78 24.2612628936768
79 24.2652778625488
80 24.4533138275146
81 23.9025192260742
82 24.303352355957
83 24.1576156616211
84 24.2483901977539
85 23.9774913787842
86 24.1850357055664
87 24.0435905456543
88 24.5996551513672
89 24.3408260345459
90 24.3080215454102
91 24.1403369903564
92 24.493782043457
93 23.8753967285156
94 24.308666229248
95 24.3791961669922
96 24.2496376037598
97 24.0878448486328
98 24.332935333252
99 24.1508483886719
100 24.1408004760742
101 24.3325881958008
102 23.8855247497559
103 23.9666996002197
104 24.1041603088379
105 23.7139320373535
106 24.3576278686523
107 24.5743541717529
108 23.9405765533447
109 24.1792182922363
110 24.1609554290771
111 24.3272247314453
112 24.1369705200195
113 24.5590209960938
114 23.5156784057617
115 24.2698192596436
116 23.7741203308105
117 24.4518737792969
118 24.01708984375
119 24.4433059692383
120 24.1472625732422
121 23.8213882446289
122 23.8043766021729
123 24.249979019165
124 23.891529083252
125 24.3051300048828
126 23.9457931518555
127 23.9912433624268
128 24.1668796539307
129 24.0996589660645
130 24.1811943054199
131 23.8041343688965
132 24.1027927398682
133 24.314582824707
134 24.1452522277832
135 23.9078540802002
136 24.2622852325439
137 24.1948013305664
138 24.3156852722168
139 24.3264503479004
140 23.8798065185547
141 24.2762966156006
142 24.1743717193604
143 24.0080375671387
144 24.5698337554932
145 24.5271530151367
146 24.4195957183838
147 24.3404788970947
148 24.2264251708984
149 24.1544647216797
150 24.5672454833984
151 24.1540660858154
152 24.0407733917236
153 23.836856842041
154 24.0218982696533
155 24.0401496887207
156 24.1627941131592
157 23.9154586791992
158 24.2726192474365
159 23.565954208374
160 24.0476417541504
161 23.6195030212402
162 24.1128311157227
163 24.3084411621094
164 24.5804443359375
165 23.7799530029297
166 24.5256042480469
167 23.7773742675781
168 23.6423568725586
169 23.940299987793
170 23.8854084014893
171 23.6644420623779
172 24.103946685791
173 24.3046531677246
174 23.9456596374512
175 23.8776588439941
176 23.4315910339355
177 24.0693168640137
178 23.3514175415039
179 24.3550224304199
180 23.7222747802734
181 24.1374206542969
182 24.0066623687744
183 23.807336807251
184 24.2296924591064
185 23.8457889556885
186 23.9782543182373
187 23.8389339447021
188 24.2107334136963
189 24.3174610137939
190 24.1990642547607
191 24.010856628418
192 24.3800239562988
193 24.3149299621582
194 23.8281326293945
195 24.3163776397705
196 23.7853927612305
197 23.8065567016602
198 23.6354103088379
199 24.3312644958496
};
\addplot [very thick, royalblue]
table {%
0 24.2336654663086
1 23.8404426574707
2 24.4670505523682
3 23.8270206451416
4 24.1517314910889
5 23.6698608398438
6 23.7781410217285
7 23.8336029052734
8 24.5533466339111
9 24.2795524597168
10 23.8403339385986
11 24.2953453063965
12 24.1232414245605
13 24.1096954345703
14 24.1460952758789
15 24.2736263275146
16 23.934549331665
17 23.8408012390137
18 24.0154781341553
19 23.8960399627686
20 24.009822845459
21 24.2718830108643
22 24.2460136413574
23 24.1309700012207
24 24.2785224914551
25 24.1430473327637
26 23.3293113708496
27 24.4460391998291
28 24.1900215148926
29 24.2146835327148
30 24.3423080444336
31 23.7753009796143
32 24.1706123352051
33 23.8102264404297
34 24.0641632080078
35 23.8836898803711
36 23.7698516845703
37 23.8667812347412
38 24.5472164154053
39 24.1872863769531
40 23.7790794372559
41 23.9464340209961
42 23.6088256835938
43 24.420352935791
44 24.0284080505371
45 23.437162399292
46 23.6685485839844
47 24.1327209472656
48 24.1343650817871
49 23.9788265228271
50 24.0013694763184
51 24.2073211669922
52 24.2439804077148
53 24.0156345367432
54 24.1892051696777
55 23.3855266571045
56 24.5460529327393
57 24.0738945007324
58 23.9704608917236
59 23.9746627807617
60 24.3997783660889
61 24.4584407806396
62 23.7780456542969
63 24.2528457641602
64 24.0702972412109
65 24.0424880981445
66 23.9833316802979
67 24.0257053375244
68 24.1411151885986
69 23.8339996337891
70 24.0256366729736
71 24.6232414245605
72 23.5338439941406
73 23.9950771331787
74 23.8591575622559
75 24.3092708587646
76 24.0621585845947
77 24.3329238891602
78 24.2612628936768
79 24.2652778625488
80 24.4533138275146
81 23.9025192260742
82 24.303352355957
83 24.1576156616211
84 24.2483901977539
85 23.9774913787842
86 24.1850357055664
87 24.0435905456543
88 24.5996551513672
89 24.3408260345459
90 24.3080215454102
91 24.1539897918701
92 24.4931488037109
93 23.7768650054932
94 24.3349685668945
95 24.3860931396484
96 24.2336750030518
97 24.1404190063477
98 24.3505744934082
99 24.2016410827637
100 24.1047248840332
101 24.3351440429688
102 23.7841453552246
103 23.9966449737549
104 23.932689666748
105 23.7786083221436
106 24.2593688964844
107 24.5502853393555
108 23.9617919921875
109 24.0561752319336
110 24.2032413482666
111 24.2580223083496
112 24.1848430633545
113 24.5490303039551
114 23.5592384338379
115 24.2642974853516
116 23.7903709411621
117 24.343729019165
118 24.1631393432617
119 24.3981285095215
120 24.2485847473145
121 23.8711566925049
122 23.826000213623
123 24.2815246582031
124 23.7762565612793
125 24.2658824920654
126 24.0486297607422
127 24.0294399261475
128 23.8373260498047
129 23.7993316650391
130 24.3077850341797
131 23.8035736083984
132 24.1302719116211
133 24.5403385162354
134 24.112886428833
135 23.8249988555908
136 24.2684783935547
137 24.0871849060059
138 24.3103866577148
139 24.2817554473877
140 23.7997932434082
141 24.3037872314453
142 24.1502342224121
143 24.1852893829346
144 24.4743347167969
145 24.5772476196289
146 24.4154663085938
147 24.2239246368408
148 24.2147064208984
149 24.1932220458984
150 24.494909286499
151 24.1816139221191
152 24.1330604553223
153 23.7633438110352
154 23.9260520935059
155 24.0492172241211
156 24.153392791748
157 23.8385543823242
158 24.1690711975098
159 23.5842895507812
160 23.9483814239502
161 23.4784355163574
162 24.1079616546631
163 24.3303508758545
164 24.4580898284912
165 23.6974296569824
166 24.4405326843262
167 23.7029857635498
168 23.6685943603516
169 23.9187545776367
170 23.836742401123
171 23.7286548614502
172 24.1988964080811
173 24.2529563903809
174 23.7799968719482
175 23.6978721618652
176 23.4712142944336
177 23.9652557373047
178 23.4707946777344
179 24.3712825775146
180 23.8025169372559
181 24.2566394805908
182 23.929084777832
183 23.7908306121826
184 24.2721061706543
185 23.8118667602539
186 23.8064727783203
187 23.6985168457031
188 24.285228729248
189 24.2381439208984
190 24.3200416564941
191 23.8480625152588
192 24.3453617095947
193 24.1870441436768
194 23.8899993896484
195 24.244800567627
196 23.7908096313477
197 23.8101806640625
198 23.62233543396
199 24.4245777130127
};
\end{axis}

\end{tikzpicture}
%\caption{}
\label{fig:energy_td_proposed}
\end{subfigure}\hspace{2em}
\begin{subfigure}[b]{0.30\textwidth}
\centering
% This file was created with tikzplotlib v0.10.1.
\begin{tikzpicture}[scale=0.6]
\pgfplotsset{
every axis legend/.append style={at={(0.5,1.1)}, anchor=center,legend columns = 2},
legend style={/tikz/every even column/.append style={column sep=0.5cm}}}

\definecolor{darkgray176}{RGB}{176,176,176}
\definecolor{darkorange25512714}{RGB}{255,127,14}
\definecolor{steelblue31119180}{RGB}{31,119,180}
\definecolor{crimson}{RGB}{220,20,60}

\begin{axis}[
tick align=outside,
tick pos=left,
x grid style={darkgray176},
xlabel={\textbf{Portion of Test Data Samples}},
xmajorgrids,
xmin=-9.95, xmax=208.95,
xtick style={color=black},
y grid style={darkgray176},
ylabel={\textbf{Indoor Temperature}},
ymajorgrids,
ymin=23.2621487617493, ymax=24.7397261619568,
ytick style={color=black}
]
\addlegendimage{area legend,pattern=north west lines, color=darkorange25512714,draw=white}
\addlegendimage{area legend,pattern=north west lines, color=crimson,draw=white}
\addlegendentry{Actual}
\addlegendentry{Predicted (GU)}

\addplot [very thick, darkorange25512714]
table {%
0 24.307580947876
1 23.7987747192383
2 24.4434185028076
3 24.0948219299316
4 24.1505222320557
5 23.5697498321533
6 23.8004264831543
7 23.9428977966309
8 24.6725635528564
9 24.2956047058105
10 23.8504047393799
11 24.2559108734131
12 24.124116897583
13 24.0542602539062
14 24.224328994751
15 24.2713832855225
16 24.0202140808105
17 23.5954093933105
18 24.000301361084
19 23.9467964172363
20 24.009822845459
21 24.2718830108643
22 24.2460136413574
23 24.1309700012207
24 24.2785224914551
25 24.1430473327637
26 23.3293113708496
27 24.4460391998291
28 24.1900215148926
29 24.2146835327148
30 24.3423080444336
31 23.7450923919678
32 24.2156562805176
33 23.7900829315186
34 24.0960960388184
35 23.9963130950928
36 23.8544940948486
37 23.8570175170898
38 24.3753395080566
39 24.1999740600586
40 23.7790794372559
41 23.9464340209961
42 23.6088256835938
43 24.420352935791
44 24.0284080505371
45 23.437162399292
46 23.6685485839844
47 24.1327209472656
48 24.1343650817871
49 23.9788265228271
50 24.0013694763184
51 24.2073211669922
52 24.2439804077148
53 24.0156345367432
54 24.1892051696777
55 23.3855266571045
56 24.5460529327393
57 24.0738945007324
58 23.9704608917236
59 23.9746627807617
60 24.3997783660889
61 24.4584407806396
62 23.7780456542969
63 24.2528457641602
64 24.0702972412109
65 24.0424880981445
66 23.9833316802979
67 24.0257053375244
68 24.1411151885986
69 23.8339996337891
70 24.0256366729736
71 24.6232414245605
72 23.5338439941406
73 23.9950771331787
74 23.8591575622559
75 24.3092708587646
76 24.0621585845947
77 24.3329238891602
78 24.2612628936768
79 24.2652778625488
80 24.4533138275146
81 23.9025192260742
82 24.303352355957
83 24.1576156616211
84 24.2483901977539
85 23.9774913787842
86 24.1850357055664
87 24.0435905456543
88 24.5996551513672
89 24.3408260345459
90 24.3080215454102
91 24.1403369903564
92 24.493782043457
93 23.8753967285156
94 24.308666229248
95 24.3791961669922
96 24.2496376037598
97 24.0878448486328
98 24.332935333252
99 24.1508483886719
100 24.1408004760742
101 24.3325881958008
102 23.8855247497559
103 23.9666996002197
104 24.1041603088379
105 23.7139320373535
106 24.3576278686523
107 24.5743541717529
108 23.9405765533447
109 24.1792182922363
110 24.1609554290771
111 24.3272247314453
112 24.1369705200195
113 24.5590209960938
114 23.5156784057617
115 24.2698192596436
116 23.7741203308105
117 24.4518737792969
118 24.01708984375
119 24.4433059692383
120 24.1472625732422
121 23.8213882446289
122 23.8043766021729
123 24.249979019165
124 23.891529083252
125 24.3051300048828
126 23.9457931518555
127 23.9912433624268
128 24.1668796539307
129 24.0996589660645
130 24.1811943054199
131 23.8041343688965
132 24.1027927398682
133 24.314582824707
134 24.1452522277832
135 23.9078540802002
136 24.2622852325439
137 24.1948013305664
138 24.3156852722168
139 24.3264503479004
140 23.8798065185547
141 24.2762966156006
142 24.1743717193604
143 24.0080375671387
144 24.5698337554932
145 24.5271530151367
146 24.4195957183838
147 24.3404788970947
148 24.2264251708984
149 24.1544647216797
150 24.5672454833984
151 24.1540660858154
152 24.0407733917236
153 23.836856842041
154 24.0218982696533
155 24.0401496887207
156 24.1627941131592
157 23.9154586791992
158 24.2726192474365
159 23.565954208374
160 24.0476417541504
161 23.6195030212402
162 24.1128311157227
163 24.3084411621094
164 24.5804443359375
165 23.7799530029297
166 24.5256042480469
167 23.7773742675781
168 23.6423568725586
169 23.940299987793
170 23.8854084014893
171 23.6644420623779
172 24.103946685791
173 24.3046531677246
174 23.9456596374512
175 23.8776588439941
176 23.4315910339355
177 24.0693168640137
178 23.3514175415039
179 24.3550224304199
180 23.7222747802734
181 24.1374206542969
182 24.0066623687744
183 23.807336807251
184 24.2296924591064
185 23.8457889556885
186 23.9782543182373
187 23.8389339447021
188 24.2107334136963
189 24.3174610137939
190 24.1990642547607
191 24.010856628418
192 24.3800239562988
193 24.3149299621582
194 23.8281326293945
195 24.3163776397705
196 23.7853927612305
197 23.8065567016602
198 23.6354103088379
199 24.3312644958496
};
\addplot [very thick, crimson]
table {%
0 24.2793865203857
1 24.0853805541992
2 24.4188060760498
3 24.1022396087646
4 24.1664905548096
5 24.1087837219238
6 24.1057739257812
7 24.0953216552734
8 24.6377983093262
9 24.3374423980713
10 24.0956840515137
11 24.3271522521973
12 24.1488227844238
13 24.1413040161133
14 24.1671810150146
15 24.3178901672363
16 24.073221206665
17 24.0838050842285
18 24.0898361206055
19 24.0990905761719
20 24.0985832214355
21 24.2116622924805
22 24.2184028625488
23 24.086555480957
24 24.371883392334
25 24.3438606262207
26 24.1155586242676
27 24.1712951660156
28 24.3242568969727
29 24.1328792572021
30 24.0965023040771
31 24.1068572998047
32 24.2270698547363
33 24.0921859741211
34 24.1112403869629
35 24.0989799499512
36 24.1038932800293
37 24.1001167297363
38 24.4074401855469
39 24.1979293823242
40 24.1078033447266
41 24.0870532989502
42 24.1062145233154
43 24.179557800293
44 24.144645690918
45 24.1113471984863
46 24.10888671875
47 24.1218452453613
48 24.151725769043
49 24.1584815979004
50 24.079496383667
51 24.3511028289795
52 24.3405952453613
53 24.1008453369141
54 24.1880264282227
55 24.1153335571289
56 24.4855613708496
57 24.1664848327637
58 24.0848388671875
59 24.0991401672363
60 24.398193359375
61 24.493579864502
62 24.101245880127
63 24.3608818054199
64 24.1818695068359
65 24.1025466918945
66 24.0844955444336
67 24.1286354064941
68 24.3751373291016
69 24.1072940826416
70 24.0951671600342
71 24.3924160003662
72 24.1097755432129
73 24.0702838897705
74 24.0829772949219
75 24.4025173187256
76 24.2356090545654
77 24.3934936523438
78 24.1923599243164
79 24.3170623779297
80 24.5834941864014
81 24.0666961669922
82 24.3953914642334
83 24.1413707733154
84 24.3612403869629
85 24.0966053009033
86 24.3243522644043
87 24.3603191375732
88 24.6544189453125
89 24.3320236206055
90 24.3778800964355
91 24.1697902679443
92 24.4743175506592
93 24.1045036315918
94 24.3535614013672
95 24.3951377868652
96 24.293571472168
97 24.1809692382812
98 24.3782768249512
99 24.2376937866211
100 24.148509979248
101 24.3868770599365
102 24.1050262451172
103 24.0909767150879
104 24.0811805725098
105 24.0994873046875
106 24.2969551086426
107 24.60032081604
108 24.1010437011719
109 24.1178188323975
110 24.2252006530762
111 24.3038349151611
112 24.2406387329102
113 24.5909767150879
114 24.1137599945068
115 24.236873626709
116 24.1023292541504
117 24.2773933410645
118 24.0886268615723
119 24.2230052947998
120 24.3127403259277
121 24.0791664123535
122 24.0962505340576
123 24.324369430542
124 24.1035976409912
125 24.2275714874268
126 24.0944576263428
127 24.0789909362793
128 24.0875797271729
129 24.0935363769531
130 24.3428745269775
131 24.1036071777344
132 24.146110534668
133 24.4278144836426
134 24.132682800293
135 24.0982971191406
136 24.3289375305176
137 24.1167030334473
138 24.3750762939453
139 24.3458938598633
140 24.1021823883057
141 24.2939395904541
142 24.1727542877197
143 24.1998424530029
144 24.4636211395264
145 24.658374786377
146 24.4301452636719
147 24.2565040588379
148 24.2454299926758
149 24.2319297790527
150 24.536283493042
151 24.215784072876
152 24.1675186157227
153 24.102222442627
154 24.0851993560791
155 24.1134433746338
156 24.1892948150635
157 24.0992183685303
158 24.1821556091309
159 24.1147613525391
160 24.0761699676514
161 24.1146945953369
162 24.0788688659668
163 24.3835258483887
164 24.4522933959961
165 24.1069145202637
166 24.1504249572754
167 24.1110801696777
168 24.1086559295654
169 24.0699596405029
170 24.0938472747803
171 24.1079788208008
172 24.1860218048096
173 24.2815361022949
174 24.1051063537598
175 24.1106967926025
176 24.1144428253174
177 24.0938682556152
178 24.1145515441895
179 24.2987422943115
180 24.1009368896484
181 24.1912250518799
182 24.100061416626
183 24.1037788391113
184 24.358455657959
185 24.0928020477295
186 24.0994606018066
187 24.1092681884766
188 24.2307586669922
189 24.2575569152832
190 24.2943477630615
191 24.0977878570557
192 24.3877792358398
193 24.2293243408203
194 24.0911617279053
195 24.288179397583
196 24.1024608612061
197 24.1010627746582
198 24.111759185791
199 24.2771224975586
};
\end{axis}

\end{tikzpicture}
%\caption{}
\label{fig:energy_td_gu}   
\end{subfigure}
\hspace{2em}
\begin{subfigure}[b]{0.30\textwidth}
\centering
% This file was created with tikzplotlib v0.10.1.
\begin{tikzpicture}[scale=0.6]
\pgfplotsset{
every axis legend/.append style={at={(0.5,1.1)}, anchor=center,legend columns = 2},
legend style={/tikz/every even column/.append style={column sep=0.5cm}}}

\definecolor{darkslategray38}{RGB}{38,38,38}
\definecolor{lightgray204}{RGB}{204,204,204}
\definecolor{peru22113282}{RGB}{221,132,82}
\definecolor{steelblue76114176}{RGB}{76,114,176}
\definecolor{green}{RGB}{0,128,0}

\begin{axis}[
tick align=outside,
tick pos=left,
x grid style={lightgray204},
xlabel={\textbf{Portion of Test Data Samples}},
xmajorgrids,
xmin=-9.95, xmax=208.95,
xtick style={color=darkslategray38},
y grid style={lightgray204},
ylabel={\textbf{Indoor Temperature}},
ymajorgrids,
ymin=23.2621487617493, ymax=24.7397261619568,
ytick style={color=darkslategray38}
]
\addlegendimage{area legend,pattern=north west lines, color=peru22113282,draw=white}
\addlegendimage{area legend,pattern=north west lines, color=green,draw=white}
\addlegendentry{Actual}
\addlegendentry{Predicted (EWC)}

\addplot [very thick, peru22113282]
table {%
0 24.307580947876
1 23.7987747192383
2 24.4434185028076
3 24.0948219299316
4 24.1505222320557
5 23.5697498321533
6 23.8004264831543
7 23.9428977966309
8 24.6725635528564
9 24.2956047058105
10 23.8504047393799
11 24.2559108734131
12 24.124116897583
13 24.0542602539062
14 24.224328994751
15 24.2713832855225
16 24.0202140808105
17 23.5954093933105
18 24.000301361084
19 23.9467964172363
20 24.009822845459
21 24.2718830108643
22 24.2460136413574
23 24.1309700012207
24 24.2785224914551
25 24.1430473327637
26 23.3293113708496
27 24.4460391998291
28 24.1900215148926
29 24.2146835327148
30 24.3423080444336
31 23.7450923919678
32 24.2156562805176
33 23.7900829315186
34 24.0960960388184
35 23.9963130950928
36 23.8544940948486
37 23.8570175170898
38 24.3753395080566
39 24.1999740600586
40 23.7790794372559
41 23.9464340209961
42 23.6088256835938
43 24.420352935791
44 24.0284080505371
45 23.437162399292
46 23.6685485839844
47 24.1327209472656
48 24.1343650817871
49 23.9788265228271
50 24.0013694763184
51 24.2073211669922
52 24.2439804077148
53 24.0156345367432
54 24.1892051696777
55 23.3855266571045
56 24.5460529327393
57 24.0738945007324
58 23.9704608917236
59 23.9746627807617
60 24.3997783660889
61 24.4584407806396
62 23.7780456542969
63 24.2528457641602
64 24.0702972412109
65 24.0424880981445
66 23.9833316802979
67 24.0257053375244
68 24.1411151885986
69 23.8339996337891
70 24.0256366729736
71 24.6232414245605
72 23.5338439941406
73 23.9950771331787
74 23.8591575622559
75 24.3092708587646
76 24.0621585845947
77 24.3329238891602
78 24.2612628936768
79 24.2652778625488
80 24.4533138275146
81 23.9025192260742
82 24.303352355957
83 24.1576156616211
84 24.2483901977539
85 23.9774913787842
86 24.1850357055664
87 24.0435905456543
88 24.5996551513672
89 24.3408260345459
90 24.3080215454102
91 24.1403369903564
92 24.493782043457
93 23.8753967285156
94 24.308666229248
95 24.3791961669922
96 24.2496376037598
97 24.0878448486328
98 24.332935333252
99 24.1508483886719
100 24.1408004760742
101 24.3325881958008
102 23.8855247497559
103 23.9666996002197
104 24.1041603088379
105 23.7139320373535
106 24.3576278686523
107 24.5743541717529
108 23.9405765533447
109 24.1792182922363
110 24.1609554290771
111 24.3272247314453
112 24.1369705200195
113 24.5590209960938
114 23.5156784057617
115 24.2698192596436
116 23.7741203308105
117 24.4518737792969
118 24.01708984375
119 24.4433059692383
120 24.1472625732422
121 23.8213882446289
122 23.8043766021729
123 24.249979019165
124 23.891529083252
125 24.3051300048828
126 23.9457931518555
127 23.9912433624268
128 24.1668796539307
129 24.0996589660645
130 24.1811943054199
131 23.8041343688965
132 24.1027927398682
133 24.314582824707
134 24.1452522277832
135 23.9078540802002
136 24.2622852325439
137 24.1948013305664
138 24.3156852722168
139 24.3264503479004
140 23.8798065185547
141 24.2762966156006
142 24.1743717193604
143 24.0080375671387
144 24.5698337554932
145 24.5271530151367
146 24.4195957183838
147 24.3404788970947
148 24.2264251708984
149 24.1544647216797
150 24.5672454833984
151 24.1540660858154
152 24.0407733917236
153 23.836856842041
154 24.0218982696533
155 24.0401496887207
156 24.1627941131592
157 23.9154586791992
158 24.2726192474365
159 23.565954208374
160 24.0476417541504
161 23.6195030212402
162 24.1128311157227
163 24.3084411621094
164 24.5804443359375
165 23.7799530029297
166 24.5256042480469
167 23.7773742675781
168 23.6423568725586
169 23.940299987793
170 23.8854084014893
171 23.6644420623779
172 24.103946685791
173 24.3046531677246
174 23.9456596374512
175 23.8776588439941
176 23.4315910339355
177 24.0693168640137
178 23.3514175415039
179 24.3550224304199
180 23.7222747802734
181 24.1374206542969
182 24.0066623687744
183 23.807336807251
184 24.2296924591064
185 23.8457889556885
186 23.9782543182373
187 23.8389339447021
188 24.2107334136963
189 24.3174610137939
190 24.1990642547607
191 24.010856628418
192 24.3800239562988
193 24.3149299621582
194 23.8281326293945
195 24.3163776397705
196 23.7853927612305
197 23.8065567016602
198 23.6354103088379
199 24.3312644958496
};
\addplot [very thick, green]
table {%
0 24.1636180877686
1 23.9101448059082
2 24.4032173156738
3 23.9439811706543
4 24.1346492767334
5 23.7935199737549
6 23.8118648529053
7 23.8503589630127
8 24.5066051483154
9 24.2777462005615
10 23.8898983001709
11 24.2611045837402
12 24.1416702270508
13 24.1212539672852
14 24.1216926574707
15 24.2168636322021
16 23.9921073913574
17 23.899730682373
18 23.9991989135742
19 23.9984188079834
20 24.0641689300537
21 24.165843963623
22 24.1519088745117
23 24.0402374267578
24 24.185173034668
25 24.2379055023193
26 23.5548572540283
27 24.2272758483887
28 24.2861461639404
29 24.2120666503906
30 24.3803558349609
31 23.8674392700195
32 24.1183052062988
33 23.8815212249756
34 24.0385646820068
35 23.9923496246338
36 23.8867816925049
37 23.9653739929199
38 24.4940872192383
39 24.1551513671875
40 23.8141765594482
41 23.84547996521
42 23.8284244537354
43 24.1262493133545
44 24.1252632141113
45 23.669734954834
46 23.7064990997314
47 23.9707450866699
48 24.0885047912598
49 24.1035785675049
50 23.9780540466309
51 24.2503261566162
52 24.2972965240479
53 24.0943546295166
54 24.1234130859375
55 23.6599254608154
56 24.4243412017822
57 24.1668853759766
58 23.9547100067139
59 24.0733642578125
60 24.3649291992188
61 24.4484615325928
62 23.9835109710693
63 24.2450485229492
64 24.1055068969727
65 24.0597019195557
66 23.9512252807617
67 24.1084442138672
68 24.3042964935303
69 23.8318862915039
70 23.912181854248
71 24.3351287841797
72 23.7825126647949
73 23.8468132019043
74 23.9445190429688
75 24.3245048522949
76 24.2125835418701
77 24.3094825744629
78 24.1540679931641
79 24.2106094360352
80 24.4740867614746
81 23.9774322509766
82 24.3537731170654
83 24.1352233886719
84 24.3107681274414
85 24.0595054626465
86 24.2281093597412
87 24.2636547088623
88 24.5146522521973
89 24.3208236694336
90 24.29762840271
91 24.1604995727539
92 24.4641227722168
93 23.8864097595215
94 24.2760696411133
95 24.3373527526855
96 24.1537933349609
97 24.1445045471191
98 24.2954025268555
99 24.2053184509277
100 24.1303806304932
101 24.2953453063965
102 23.920051574707
103 23.9814739227295
104 23.9905738830566
105 23.8548316955566
106 24.2003860473633
107 24.5082130432129
108 24.0479869842529
109 24.0859909057617
110 24.1483707427979
111 24.2054538726807
112 24.1256523132324
113 24.5029621124268
114 23.6786422729492
115 24.2093067169189
116 23.9198875427246
117 24.3041496276855
118 24.145881652832
119 24.3289260864258
120 24.1634788513184
121 23.9368476867676
122 23.8855476379395
123 24.2365474700928
124 23.8926811218262
125 24.220386505127
126 24.0654411315918
127 23.9955997467041
128 23.9337501525879
129 23.8672904968262
130 24.2382259368896
131 23.920804977417
132 24.1222267150879
133 24.4121513366699
134 24.1247692108154
135 23.9594402313232
136 24.1757259368896
137 24.0949153900146
138 24.2609634399414
139 24.2706623077393
140 23.922643661499
141 24.2593536376953
142 24.1426639556885
143 24.1506690979004
144 24.4059276580811
145 24.5339736938477
146 24.4230690002441
147 24.2410545349121
148 24.2466011047363
149 24.206958770752
150 24.4803657531738
151 24.2312240600586
152 24.0944042205811
153 23.8466644287109
154 23.9279441833496
155 24.0789451599121
156 24.1614475250244
157 24.0025177001953
158 24.0956783294678
159 23.7482223510742
160 23.99440574646
161 23.6258869171143
162 23.9514083862305
163 24.2778244018555
164 24.3715896606445
165 23.8369293212891
166 24.3394012451172
167 23.7858543395996
168 23.7999782562256
169 23.8512840270996
170 23.8874092102051
171 23.826358795166
172 24.1409034729004
173 24.2124824523926
174 23.9351577758789
175 23.8419189453125
176 23.6135139465332
177 23.7424430847168
178 23.6292057037354
179 24.2900810241699
180 23.9468116760254
181 24.1666889190674
182 24.0020065307617
183 23.8817863464355
184 24.177282333374
185 23.8765468597412
186 23.8707180023193
187 23.7908592224121
188 24.2320861816406
189 24.228588104248
190 24.2558403015137
191 23.948902130127
192 24.2379665374756
193 24.1899185180664
194 23.9217739105225
195 24.2187919616699
196 23.9172973632812
197 23.9490985870361
198 23.7792282104492
199 24.3181800842285
};
\end{axis}

\end{tikzpicture}
%\caption{}
\label{fig:energy_td_ewc}    
\end{subfigure}
 \caption{Performance comparison between \emph{One-step fine-tuning} and fine-tuning baselines in terms of Actual v.s. Predicted indoor temperature values of target domain NIST.}
    \label{fig:energy_avsp_td}
\end{figure*}

%% file: mavsp.tex
\begin{figure*}[t!]
\centering
\begin{subfigure}[b]{0.30\textwidth}
\centering
% This file was created with tikzplotlib v0.10.1.
\begin{tikzpicture}[scale=0.6]
\pgfplotsset{
every axis legend/.append style={at={(0.5,1.1)}, anchor=center,legend columns = 2},
legend style={/tikz/every even column/.append style={column sep=0.5cm}}}

\definecolor{darkgray176}{RGB}{176,176,176}
\definecolor{darkorange25512714}{RGB}{255,127,14}
\definecolor{steelblue31119180}{RGB}{31,119,180}
\definecolor{royalblue}{RGB}{65,105,225}

\begin{axis}[
tick align=outside,
tick pos=left,
x grid style={darkgray176},
xmajorgrids,
xlabel={\textbf{Portion of Test Data Samples}},
xmin=-9.95, xmax=208.95,
xtick style={color=black},
y grid style={darkgray176},
ymajorgrids,
ylabel={\textbf{Wind Power}},
ymin=-8.92203559875488, ymax=43.6748279571533,
ytick style={color=black}
]
\addlegendimage{area legend,pattern=north west lines, color=darkorange25512714,draw=white}
\addlegendimage{area legend,pattern=north west lines, color=royalblue,draw=white}
\addlegendentry{Actual}
\addlegendentry{Predicted (\emph{One-step fine-tuning})}

\addplot [very thick, darkorange25512714]
table {%
0 13.6143426895142
1 14.8854284286499
2 8.31053447723389
3 9.86146354675293
4 12.1473522186279
5 15.8913116455078
6 13.9303007125854
7 7.42496967315674
8 2.95995259284973
9 0.392074257135391
10 -0.126117557287216
11 0.0315504372119904
12 -0.451116383075714
13 -2.25959205627441
14 -2.79911041259766
15 -4.12945461273193
16 -4.50411891937256
17 -3.83348417282104
18 -3.27940511703491
19 1.54513704776764
20 0.908649146556854
21 -0.268281280994415
22 1.719850897789
23 -0.640529096126556
24 2.16163468360901
25 -0.335183590650558
26 0.688074946403503
27 -0.258741825819016
28 -2.66808104515076
29 -4.74645614624023
30 -6.14638185501099
31 -6.10141611099243
32 -5.37690687179565
33 -2.70596408843994
34 -2.79427075386047
35 -3.41624712944031
36 -2.52024126052856
37 -1.85454225540161
38 -1.08849895000458
39 3.40514707565308
40 3.96045637130737
41 8.50542545318604
42 13.1659126281738
43 23.4844188690186
44 19.3773193359375
45 16.7803401947021
46 14.6919384002686
47 16.3376731872559
48 3.7913281917572
49 11.4891710281372
50 9.05327129364014
51 8.22758102416992
52 8.54476737976074
53 4.80559587478638
54 2.17349028587341
55 -2.45842695236206
56 -2.59049463272095
57 -0.172962814569473
58 -4.01155042648315
59 -5.15316867828369
60 -5.26490545272827
61 -5.64989137649536
62 -5.08962440490723
63 -5.20058250427246
64 -5.79838085174561
65 -5.92541599273682
66 -5.50406789779663
67 -4.88968896865845
68 -1.30855822563171
69 13.3699703216553
70 13.4153442382812
71 14.5926570892334
72 10.0343494415283
73 9.3623218536377
74 10.5085039138794
75 12.5326118469238
76 9.66073226928711
77 10.9121541976929
78 3.47998094558716
79 11.8616590499878
80 10.7669172286987
81 14.8140096664429
82 19.1570377349854
83 19.5147171020508
84 5.6502947807312
85 1.93807947635651
86 9.97860431671143
87 16.1258087158203
88 16.8547763824463
89 16.064661026001
90 16.9630126953125
91 20.5821380615234
92 20.6446323394775
93 13.7412233352661
94 15.9312438964844
95 9.9897518157959
96 11.3401918411255
97 9.83491039276123
98 11.1260509490967
99 8.77493476867676
100 12.0824937820435
101 18.185848236084
102 21.2641944885254
103 17.3526058197021
104 15.6223621368408
105 14.0045738220215
106 22.5972785949707
107 22.6095237731934
108 29.6017723083496
109 15.8741807937622
110 31.7925930023193
111 36.7504272460938
112 38.6166305541992
113 36.8510246276855
114 38.351921081543
115 33.2760772705078
116 29.3368797302246
117 23.9037036895752
118 24.617582321167
119 33.9072799682617
120 35.3198509216309
121 23.5284690856934
122 15.9061012268066
123 12.6455144882202
124 14.9810190200806
125 29.5270938873291
126 21.9338912963867
127 20.2925872802734
128 19.0030918121338
129 16.9904251098633
130 19.8870754241943
131 18.0728988647461
132 18.1774883270264
133 14.5503215789795
134 15.2731952667236
135 20.4214458465576
136 15.8944759368896
137 11.5933713912964
138 4.81700563430786
139 -2.9410560131073
140 4.87972354888916
141 2.63145804405212
142 3.94224643707275
143 4.12755870819092
144 7.84834718704224
145 12.5895147323608
146 10.2214384078979
147 7.98072862625122
148 4.72248697280884
149 8.78242778778076
150 9.68261528015137
151 16.6744747161865
152 21.2783012390137
153 23.8528575897217
154 25.2976417541504
155 29.0994110107422
156 34.3875694274902
157 33.3011589050293
158 29.7724876403809
159 32.0946769714355
160 17.645170211792
161 17.0980052947998
162 19.2451477050781
163 17.804838180542
164 18.9973545074463
165 19.3086490631104
166 21.7714405059814
167 21.5483455657959
168 23.071704864502
169 31.2408580780029
170 30.0863933563232
171 26.7303009033203
172 36.5846099853516
173 36.5226669311523
174 34.2730903625488
175 33.7114562988281
176 30.8878116607666
177 33.2768783569336
178 34.9254684448242
179 35.1295623779297
180 36.3123817443848
181 39.1772575378418
182 40.6046333312988
183 40.1380805969238
184 41.2840614318848
185 41.1127967834473
186 40.178638458252
187 39.5894927978516
188 37.8054580688477
189 38.028076171875
190 37.5250282287598
191 37.6889266967773
192 37.7013130187988
193 36.9598007202148
194 34.3716697692871
195 32.8330307006836
196 35.4814605712891
197 31.8466815948486
198 29.2942085266113
199 29.4465007781982
};
\addplot [very thick, royalblue]
table {%
0 13.6143426895142
1 14.8854284286499
2 8.31053447723389
3 9.86146354675293
4 12.1473522186279
5 15.8913116455078
6 13.9303007125854
7 7.42496967315674
8 2.95995259284973
9 0.392074257135391
10 -0.126117557287216
11 -0.48367357254028
12 -0.38038074970245
13 -2.27937912940979
14 -2.18886232376099
15 -3.9664294719696
16 -1.94915878772736
17 -1.67422103881836
18 1.91352987289429
19 -0.187455356121063
20 0.908649146556854
21 -0.268281280994415
22 1.719850897789
23 -0.640529096126556
24 2.16163468360901
25 -0.335183590650558
26 0.688074946403503
27 -0.258741825819016
28 -2.66808104515076
29 -4.74645614624023
30 -6.14638185501099
31 -4.60307455062866
32 -2.3016140460968
33 -2.08529162406921
34 -3.94131422042847
35 -3.61081576347351
36 -2.37073040008545
37 -0.690553486347198
38 1.09730780124664
39 1.28499114513397
40 4.70220947265625
41 9.2013683319092
42 13.319034576416
43 23.1308116912842
44 14.4519376754761
45 18.7803401947021
46 14.6919384002686
47 16.3376731872559
48 3.7913281917572
49 11.4891710281372
50 9.05327129364014
51 7.45696830749512
52 2.32975578308105
53 1.47381973266602
54 -1.82199990749359
55 -2.76538133621216
56 -1.08787715435028
57 -4.60001993179321
58 -6.09232234954834
59 -6.52993202209473
60 -5.26490545272827
61 -5.64989137649536
62 -5.08962440490723
63 -5.20058250427246
64 -5.79838085174561
65 -5.92541599273682
66 -5.50406789779663
67 -4.88968896865845
68 -1.30855822563171
69 13.3699703216553
70 13.4153442382812
71 14.5926570892334
72 10.0343494415283
73 9.3623218536377
74 10.5085039138794
75 12.5326118469238
76 9.66073226928711
77 10.9121541976929
78 3.47998094558716
79 11.8616590499878
80 10.7669172286987
81 14.8140096664429
82 19.1570377349854
83 19.5147171020508
84 5.6502947807312
85 1.93807947635651
86 9.97860431671143
87 16.1258087158203
88 16.8547763824463
89 16.064661026001
90 16.9630126953125
91 22.5821380615234
92 20.6446323394775
93 13.7412233352661
94 15.9312438964844
95 9.9897518157959
96 11.3401918411255
97 9.83491039276123
98 11.1260509490967
99 8.77493476867676
100 12.0824937820435
101 19.6891422271729
102 22.0347747802734
103 18.6452407836914
104 22.1576976776123
105 14.0045738220215
106 22.5972785949707
107 22.6095237731934
108 29.6017723083496
109 15.8741807937622
110 31.7925930023193
111 36.7504272460938
112 38.6166305541992
113 36.8510246276855
114 38.351921081543
115 33.2760772705078
116 29.3368797302246
117 23.9037036895752
118 24.617582321167
119 33.9072799682617
120 35.3198509216309
121 23.5284690856934
122 15.9061012268066
123 12.6455144882202
124 14.9810190200806
125 29.5270938873291
126 21.9338912963867
127 20.2925872802734
128 19.0030918121338
129 16.9904251098633
130 19.8870754241943
131 18.0728988647461
132 18.1774883270264
133 14.5503215789795
134 15.2731952667236
135 20.4214458465576
136 15.8944759368896
137 11.5933713912964
138 4.81700563430786
139 -2.9410560131073
140 4.87972354888916
141 2.63145804405212
142 3.94224643707275
143 4.12755870819092
144 7.84834718704224
145 12.5895147323608
146 10.2214384078979
147 7.98072862625122
148 4.72248697280884
149 8.78242778778076
150 9.68261528015137
151 14.8019466400146
152 21.1941299438477
153 24.7836494445801
154 26.6569328308105
155 29.0994110107422
156 34.3875694274902
157 33.3011589050293
158 29.7724876403809
159 32.0946769714355
160 19.9434223175049
161 17.8705024719238
162 21.9872741699219
163 22.0087947845459
164 22.8407669067383
165 22.5035552978516
166 22.5729789733887
167 22.8499412536621
168 22.9100532531738
169 21.8829898834229
170 32.2020320892334
171 24.6779518127441
172 38.4881172180176
173 35.6528205871582
174 34.396312713623
175 32.9957618713379
176 30.4117813110352
177 34.7422790527344
178 35.5093116760254
179 36.4021987915039
180 36.3123817443848
181 39.1772575378418
182 40.6046333312988
183 40.1380805969238
184 41.2840614318848
185 41.1127967834473
186 40.178638458252
187 39.5894927978516
188 37.8054580688477
189 38.028076171875
190 37.5250282287598
191 37.6889266967773
192 37.7013130187988
193 36.9598007202148
194 34.3716697692871
195 32.8330307006836
196 35.4814605712891
197 31.8466815948486
198 29.2942085266113
199 29.9465007781982
};
\end{axis}

\end{tikzpicture}

%\caption{}
\label{fig:more_td_proposed}
\end{subfigure}\hspace{2em}
\begin{subfigure}[b]{0.30\textwidth}
\centering
% This file was created with tikzplotlib v0.10.1.
\begin{tikzpicture}[scale=0.6]
\pgfplotsset{
every axis legend/.append style={at={(0.5,1.1)}, anchor=center,legend columns = 2},
legend style={/tikz/every even column/.append style={column sep=0.5cm}}}

\definecolor{darkslategray38}{RGB}{38,38,38}
\definecolor{lightgray204}{RGB}{204,204,204}
\definecolor{peru22113282}{RGB}{221,132,82}
\definecolor{steelblue76114176}{RGB}{76,114,176}
\definecolor{crimson}{RGB}{220,20,60}

\begin{axis}[
tick align=outside,
tick pos=left,
x grid style={lightgray204},
xlabel={\textbf{Portion of Test Data Samples}},
xmajorgrids,
xmin=-9.95, xmax=208.95,
xtick style={color=darkslategray38},
y grid style={lightgray204},
ylabel={\textbf{Wind Power}},
ymajorgrids,
ymin=-8.78329811096191, ymax=40.761340713501,
ytick style={color=darkslategray38}
]
\addlegendimage{area legend,pattern=north west lines, color=peru22113282,draw=white}
\addlegendimage{area legend,pattern=north west lines, color=crimson,draw=white}
\addlegendentry{Actual}
\addlegendentry{Predicted (GU)}

\addplot [very thick, peru22113282]
table {%
0 8.5063304901123
1 12.0684232711792
2 11.4314861297607
3 12.3820276260376
4 12.4122552871704
5 13.9147415161133
6 13.4817543029785
7 10.8726987838745
8 7.14932584762573
9 4.34469270706177
10 3.11211371421814
11 0.830443680286407
12 -0.124219536781311
13 0.282350987195969
14 2.63822221755981
15 1.82062494754791
16 2.3873393535614
17 2.34108209609985
18 1.92114734649658
19 7.20136117935181
20 5.10670566558838
21 6.27577018737793
22 9.77075958251953
23 8.39681148529053
24 2.89600396156311
25 2.23423099517822
26 3.66060614585876
27 2.68805742263794
28 -0.635159969329834
29 -2.23073387145996
30 -3.36599636077881
31 -3.78705954551697
32 -2.96377897262573
33 -0.541339218616486
34 -1.29862606525421
35 -2.65920352935791
36 -1.77502989768982
37 -1.19528222084045
38 1.08534324169159
39 3.62081813812256
40 4.58679485321045
41 9.83191585540771
42 13.6607503890991
43 20.6117820739746
44 17.8271141052246
45 15.5926847457886
46 14.5667858123779
47 16.5238132476807
48 6.15918827056885
49 11.5820503234863
50 12.0378351211548
51 12.4274482727051
52 10.7198610305786
53 7.83794164657593
54 7.24963617324829
55 2.15246653556824
56 0.244071155786514
57 0.916039705276489
58 -2.8690972328186
59 -4.10755252838135
60 -3.49655985832214
61 -3.31507039070129
62 -3.71178340911865
63 -3.33773422241211
64 -1.56701612472534
65 -1.11038196086884
66 -1.55268049240112
67 -0.854360342025757
68 2.66944527626038
69 16.7830257415771
70 16.5298805236816
71 17.2580261230469
72 8.14445209503174
73 11.7902555465698
74 12.4725103378296
75 13.6936359405518
76 13.7431144714355
77 14.4439001083374
78 9.49290084838867
79 10.6532831192017
80 11.3637647628784
81 12.704888343811
82 13.4548788070679
83 14.6036882400513
84 7.07865428924561
85 6.00755071640015
86 12.8277168273926
87 14.4741258621216
88 14.2773847579956
89 14.7595186233521
90 16.4759960174561
91 19.4350452423096
92 18.9402885437012
93 16.63623046875
94 17.2443962097168
95 14.4875526428223
96 8.23467922210693
97 9.89604568481445
98 11.2008752822876
99 11.2244892120361
100 12.2892942428589
101 12.1964864730835
102 14.8008899688721
103 14.2088088989258
104 13.7598237991333
105 12.4883432388306
106 15.8254537582397
107 16.4398365020752
108 21.6294708251953
109 14.3004541397095
110 20.8921051025391
111 23.5486869812012
112 26.3797588348389
113 28.4290523529053
114 30.2750205993652
115 27.0790939331055
116 24.9707164764404
117 23.0247745513916
118 23.0845375061035
119 27.7037181854248
120 24.0497035980225
121 15.1328601837158
122 12.5794906616211
123 12.9049997329712
124 13.6905813217163
125 19.3205833435059
126 14.9659605026245
127 16.0441493988037
128 15.5909519195557
129 13.8583726882935
130 14.7794733047485
131 13.8425054550171
132 13.7624444961548
133 13.6030654907227
134 11.7153215408325
135 13.371129989624
136 12.2995948791504
137 12.0101900100708
138 5.87640047073364
139 1.05260944366455
140 4.39762878417969
141 9.07005786895752
142 11.4282875061035
143 12.5774984359741
144 5.63786888122559
145 9.9111213684082
146 10.6883125305176
147 11.1736135482788
148 9.97029685974121
149 13.8549509048462
150 13.012487411499
151 13.0289297103882
152 17.6676177978516
153 18.700267791748
154 18.5423851013184
155 21.6699771881104
156 26.6924610137939
157 26.5228843688965
158 24.1272754669189
159 24.134822845459
160 15.1631193161011
161 16.1700401306152
162 18.0812339782715
163 16.984884262085
164 16.3879318237305
165 18.4092350006104
166 20.2248229980469
167 19.8447875976562
168 13.0450439453125
169 21.4945697784424
170 23.2947673797607
171 20.8250331878662
172 27.2184391021729
173 29.8884334564209
174 29.4867687225342
175 29.4073791503906
176 26.1767692565918
177 27.5998477935791
178 30.0412483215332
179 29.7423877716064
180 29.7575759887695
181 29.877758026123
182 30.6043186187744
183 31.3412189483643
184 32.0846633911133
185 32.2185401916504
186 32.1302185058594
187 32.3786849975586
188 31.9574165344238
189 32.2174835205078
190 31.366283416748
191 30.8908004760742
192 27.6386051177979
193 29.1454925537109
194 27.6758937835693
195 27.6675395965576
196 28.5277786254883
197 22.9238510131836
198 21.2654590606689
199 24.3190212249756
};
\addplot [very thick, crimson]
table {%
0 14.2435464859009
1 7.87949562072754
2 9.1770486831665
3 10.5127906799316
4 10.349928855896
5 10.4940795898438
6 4.9638147354126
7 2.29958891868591
8 0.280321359634399
9 1.80393624305725
10 0.988859534263611
11 -1.48367357254028
12 -1.38038074970245
13 -1.27937912940979
14 -3.18886232376099
15 -2.9664294719696
16 -1.94915878772736
17 -1.67422103881836
18 1.91352987289429
19 -0.187455356121063
20 1.56508374214172
21 3.43256258964539
22 0.865710258483887
23 0.4145447909832
24 -0.286165446043015
25 0.18027463555336
26 -2.70046544075012
27 -4.33997440338135
28 -5.86893510818481
29 -6.33403921127319
30 -6.42721223831177
31 -4.60307455062866
32 -2.3016140460968
33 -2.08529162406921
34 -3.94131422042847
35 -3.61081576347351
36 -2.37073040008545
37 -0.690553486347198
38 1.09730780124664
39 1.28499114513397
40 8.70220947265625
41 12.2013683319092
42 17.319034576416
43 13.1308116912842
44 15.4519376754761
45 14.8411550521851
46 10.1641540527344
47 7.85734701156616
48 7.37028694152832
49 7.21181583404541
50 8.94106101989746
51 7.45696830749512
52 2.32975578308105
53 1.47381973266602
54 -1.82199990749359
55 -2.76538133621216
56 -1.08787715435028
57 -4.60001993179321
58 -6.09232234954834
59 -6.52993202209473
60 -6.24888467788696
61 -6.18931484222412
62 -6.2842059135437
63 -6.25613927841187
64 -6.53126907348633
65 -6.21222591400146
66 -4.59085559844971
67 -0.493323862552643
68 9.08597660064697
69 9.0136137008667
70 8.70545482635498
71 9.22726440429688
72 12.5343494415283
73 14.6099405288696
74 14.4640703201294
75 12.0560712814331
76 10.0089273452759
77 2.56135272979736
78 4.49011898040771
79 7.50947380065918
80 10.8608541488647
81 15.3736572265625
82 16.1301193237305
83 6.25812387466431
84 2.02655982971191
85 9.35251426696777
86 15.1351871490479
87 15.1286945343018
88 15.9934129714966
89 16.8989925384521
90 19.5284671783447
91 17.7218952178955
92 11.6360263824463
93 14.8705577850342
94 6.3537802696228
95 14.5427331924438
96 8.60292530059814
97 10.2831020355225
98 7.50699234008789
99 9.95565986633301
100 12.1001758575439
101 15.688117980957
102 13.9573440551758
103 13.3223133087158
104 7.90947008132935
105 19.8679389953613
106 20.7882194519043
107 28.4469604492188
108 19.1385898590088
109 27.8585166931152
110 31.116340637207
111 30.4137210845947
112 30.8228816986084
113 33.9625205993652
114 29.3074798583984
115 29.341272354126
116 23.9164009094238
117 23.9198379516602
118 28.8429470062256
119 32.7533683776855
120 22.4139804840088
121 14.113715171814
122 12.2489099502563
123 12.200795173645
124 24.8328609466553
125 14.6979579925537
126 15.6449670791626
127 16.8907794952393
128 12.286714553833
129 18.0695762634277
130 16.8944072723389
131 17.2021865844727
132 16.2542209625244
133 15.5128440856934
134 16.423957824707
135 14.7653560638428
136 9.12778949737549
137 3.81709361076355
138 -1.78400528430939
139 2.508465051651
140 3.92935991287231
141 4.6986141204834
142 5.58261585235596
143 11.5189867019653
144 12.0770740509033
145 10.6716432571411
146 7.85333728790283
147 6.43435096740723
148 7.20303249359131
149 5.54920434951782
150 9.69427680969238
151 22.0233402252197
152 24.5361557006836
153 21.5977592468262
154 26.1390056610107
155 35.1277618408203
156 34.2512092590332
157 29.6690979003906
158 32.0860710144043
159 19.4729080200195
160 19.5452690124512
161 19.6714725494385
162 13.0679969787598
163 14.3441667556763
164 18.2618427276611
165 22.6432857513428
166 17.9317264556885
167 24.7610721588135
168 31.0477981567383
169 32.1532783508301
170 25.3424530029297
171 32.6779518127441
172 38.4881172180176
173 35.6528205871582
174 34.396312713623
175 31.9957618713379
176 34.4117813110352
177 34.7422790527344
178 38.5093116760254
179 38.4021987915039
180 38.1291732788086
181 37.8502235412598
182 37.8486938476562
183 37.5516128540039
184 37.2772445678711
185 36.7542877197266
186 37.0196800231934
187 36.8709487915039
188 36.5793991088867
189 37.0893707275391
190 37.9901733398438
191 37.9926567077637
192 37.0303726196289
193 33.8661041259766
194 34.153263092041
195 35.5441818237305
196 27.5486392974854
197 22.9590816497803
198 26.5189552307129
199 24.8286609649658
};
\end{axis}

\end{tikzpicture}

%\caption{}
\label{fig:more_td_gu}   
\end{subfigure}
\hspace{2em}
\begin{subfigure}[b]{0.30\textwidth}
\centering
% This file was created with tikzplotlib v0.10.1.
\begin{tikzpicture}[scale=0.6]
\pgfplotsset{
every axis legend/.append style={at={(0.5,1.1)}, anchor=center,legend columns = 2},
legend style={/tikz/every even column/.append style={column sep=0.5cm}}}

\definecolor{darkgray176}{RGB}{176,176,176}
\definecolor{darkorange25512714}{RGB}{255,127,14}
\definecolor{steelblue31119180}{RGB}{31,119,180}
\definecolor{green}{RGB}{0,128,0}

\begin{axis}[
tick align=outside,
tick pos=left,
x grid style={darkgray176},
xmajorgrids,
xlabel={\textbf{Portion of Test Data Samples}},
xmin=-9.95, xmax=208.95,
xtick style={color=black},
y grid style={darkgray176},
ymajorgrids,
ylabel={\textbf{Wind Power}},
ymin=-8.78329811096191, ymax=40.761340713501,
ytick style={color=black}
]
\addlegendimage{area legend,pattern=north west lines, color=darkorange25512714,draw=white}
\addlegendimage{area legend,pattern=north west lines, color=green,draw=white}
\addlegendentry{Actual}
\addlegendentry{Predicted (EWC)}

\addplot [very thick, darkorange25512714]
table {%
0 16.7456703186035
1 16.0034446716309
2 8.54533290863037
3 12.5970439910889
4 13.2959632873535
5 12.3978109359741
6 11.6287050247192
7 6.82713508605957
8 2.80752849578857
9 0.68810510635376
10 2.45622658729553
11 0.353923976421356
12 -0.790644705295563
13 -1.54220747947693
14 -1.25325512886047
15 -1.59077632427216
16 -0.880099833011627
17 -0.0580451786518097
18 0.46409347653389
19 2.82550454139709
20 0.123962581157684
21 5.75219202041626
22 6.68874216079712
23 4.98762893676758
24 2.6516900062561
25 -0.139459222555161
26 -0.842588365077972
27 -0.959593892097473
28 -2.16407608985901
29 -4.08591556549072
30 -4.81359195709229
31 -4.9404296875
32 -3.31900691986084
33 -1.68007755279541
34 -1.92780721187592
35 -3.01644921302795
36 -2.1663510799408
37 -1.15105545520782
38 -0.492374509572983
39 -0.142419695854187
40 1.59991717338562
41 9.18918418884277
42 11.9003925323486
43 16.0211124420166
44 14.167498588562
45 15.8657751083374
46 14.23668384552
47 14.6285696029663
48 9.1251277923584
49 7.25591516494751
50 6.83633422851562
51 10.4117498397827
52 8.22857761383057
53 4.09185981750488
54 2.23694539070129
55 -0.0192430317401886
56 -1.08007669448853
57 -0.494153261184692
58 -2.88923001289368
59 -4.41788148880005
60 -4.63882637023926
61 -4.21616888046265
62 -4.21984767913818
63 -5.08162212371826
64 -4.6367712020874
65 -5.63324975967407
66 -5.1003303527832
67 -3.53562068939209
68 -0.925503551959991
69 19.5025634765625
70 16.2979640960693
71 14.713436126709
72 15.4641923904419
73 12.6483058929443
74 14.1870422363281
75 17.5732250213623
76 14.9156446456909
77 10.6886653900146
78 4.74303150177002
79 7.4439115524292
80 7.36380815505981
81 12.0998201370239
82 17.2010021209717
83 20.7938365936279
84 8.75203895568848
85 3.49260759353638
86 7.67575645446777
87 16.8325061798096
88 19.574499130249
89 18.1257286071777
90 17.4995594024658
91 16.7345180511475
92 17.5236072540283
93 20.9835319519043
94 20.04345703125
95 12.9915208816528
96 16.8744792938232
97 11.6333875656128
98 11.9385299682617
99 11.1751022338867
100 13.9650573730469
101 13.6741333007812
102 18.597282409668
103 15.5291652679443
104 13.8910531997681
105 11.1858882904053
106 22.2254467010498
107 25.3285598754883
108 30.4672164916992
109 20.3621349334717
110 29.5641193389893
111 33.2746047973633
112 33.5244369506836
113 31.7482852935791
114 30.8964977264404
115 27.6039733886719
116 28.3836631774902
117 30.4910755157471
118 29.2713527679443
119 32.5369720458984
120 35.335620880127
121 24.0382995605469
122 15.3031549453735
123 14.3705101013184
124 15.9524335861206
125 25.6945152282715
126 15.425271987915
127 16.1803092956543
128 15.1763496398926
129 12.3435697555542
130 19.6946239471436
131 19.9693851470947
132 18.0525951385498
133 17.1722564697266
134 13.9440650939941
135 15.4401340484619
136 15.6701335906982
137 11.8586797714233
138 5.84779644012451
139 0.507292091846466
140 3.31127285957336
141 8.861496925354
142 7.31433486938477
143 7.21942710876465
144 11.1382026672363
145 10.4260025024414
146 10.6330881118774
147 11.3755302429199
148 8.28662014007568
149 10.0122222900391
150 7.61083745956421
151 11.5340051651001
152 18.3881549835205
153 21.2680416107178
154 26.222900390625
155 30.778923034668
156 33.059211730957
157 31.5527400970459
158 29.8580074310303
159 31.820894241333
160 25.949821472168
161 22.7529182434082
162 19.6932926177979
163 14.1099472045898
164 13.8836946487427
165 23.3166007995605
166 24.5975532531738
167 21.1870155334473
168 25.3165283203125
169 33.0640335083008
170 33.1725006103516
171 28.8248100280762
172 34.7459869384766
173 35.8432273864746
174 32.7835960388184
175 27.745512008667
176 20.9669952392578
177 27.0117702484131
178 35.5938339233398
179 35.7763977050781
180 33.5233840942383
181 31.6010284423828
182 29.7990207672119
183 33.8811683654785
184 36.3687477111816
185 36.7147521972656
186 36.9621849060059
187 37.0353584289551
188 37.2700080871582
189 36.6522216796875
190 36.6392364501953
191 35.7513847351074
192 36.0487861633301
193 34.7330551147461
194 33.1037292480469
195 34.5303497314453
196 35.5112533569336
197 27.1555290222168
198 19.7680034637451
199 17.3977718353271
};
\addplot [very thick, green]
table {%
0 14.2435464859009
1 7.87949562072754
2 9.1770486831665
3 10.5127906799316
4 10.349928855896
5 10.4940795898438
6 4.9638147354126
7 2.29958891868591
8 0.280321359634399
9 1.80393624305725
10 0.988859534263611
11 -1.48367357254028
12 -1.38038074970245
13 -1.27937912940979
14 -3.18886232376099
15 -2.9664294719696
16 -1.94915878772736
17 -1.67422103881836
18 1.91352987289429
19 -0.187455356121063
20 1.56508374214172
21 3.43256258964539
22 0.865710258483887
23 0.4145447909832
24 -0.286165446043015
25 0.18027463555336
26 -2.70046544075012
27 -4.33997440338135
28 -5.86893510818481
29 -6.33403921127319
30 -6.42721223831177
31 -4.60307455062866
32 -2.3016140460968
33 -2.08529162406921
34 -3.94131422042847
35 -3.61081576347351
36 -2.37073040008545
37 -0.690553486347198
38 1.09730780124664
39 1.28499114513397
40 8.70220947265625
41 12.2013683319092
42 17.319034576416
43 13.1308116912842
44 15.4519376754761
45 14.8411550521851
46 10.1641540527344
47 7.85734701156616
48 7.37028694152832
49 7.21181583404541
50 8.94106101989746
51 7.45696830749512
52 2.32975578308105
53 1.47381973266602
54 -1.82199990749359
55 -2.76538133621216
56 -1.08787715435028
57 -4.60001993179321
58 -6.09232234954834
59 -6.52993202209473
60 -6.24888467788696
61 -6.18931484222412
62 -6.2842059135437
63 -6.25613927841187
64 -6.53126907348633
65 -6.21222591400146
66 -4.59085559844971
67 -0.493323862552643
68 9.08597660064697
69 9.0136137008667
70 8.70545482635498
71 9.22726440429688
72 12.5343494415283
73 14.6099405288696
74 14.4640703201294
75 12.0560712814331
76 10.0089273452759
77 2.56135272979736
78 4.49011898040771
79 7.50947380065918
80 10.8608541488647
81 15.3736572265625
82 16.1301193237305
83 6.25812387466431
84 2.02655982971191
85 9.35251426696777
86 15.1351871490479
87 15.1286945343018
88 15.9934129714966
89 16.8989925384521
90 19.5284671783447
91 17.7218952178955
92 11.6360263824463
93 14.8705577850342
94 6.3537802696228
95 14.5427331924438
96 8.60292530059814
97 10.2831020355225
98 7.50699234008789
99 9.95565986633301
100 12.1001758575439
101 15.688117980957
102 13.9573440551758
103 13.3223133087158
104 7.90947008132935
105 19.8679389953613
106 20.7882194519043
107 28.4469604492188
108 19.1385898590088
109 27.8585166931152
110 31.116340637207
111 30.4137210845947
112 30.8228816986084
113 33.9625205993652
114 29.3074798583984
115 29.341272354126
116 23.9164009094238
117 23.9198379516602
118 28.8429470062256
119 32.7533683776855
120 22.4139804840088
121 14.113715171814
122 12.2489099502563
123 12.200795173645
124 24.8328609466553
125 14.6979579925537
126 15.6449670791626
127 16.8907794952393
128 12.286714553833
129 18.0695762634277
130 16.8944072723389
131 17.2021865844727
132 16.2542209625244
133 15.5128440856934
134 16.423957824707
135 14.7653560638428
136 9.12778949737549
137 3.81709361076355
138 -1.78400528430939
139 2.508465051651
140 3.92935991287231
141 4.6986141204834
142 5.58261585235596
143 11.5189867019653
144 12.0770740509033
145 10.6716432571411
146 7.85333728790283
147 6.43435096740723
148 7.20303249359131
149 5.54920434951782
150 9.69427680969238
151 22.0233402252197
152 24.5361557006836
153 21.5977592468262
154 26.1390056610107
155 35.1277618408203
156 34.2512092590332
157 29.6690979003906
158 32.0860710144043
159 19.4729080200195
160 19.5452690124512
161 19.6714725494385
162 13.0679969787598
163 14.3441667556763
164 18.2618427276611
165 22.6432857513428
166 17.9317264556885
167 24.7610721588135
168 31.0477981567383
169 32.1532783508301
170 25.3424530029297
171 32.6779518127441
172 38.4881172180176
173 35.6528205871582
174 34.396312713623
175 31.9957618713379
176 34.4117813110352
177 34.7422790527344
178 38.5093116760254
179 38.4021987915039
180 38.1291732788086
181 37.8502235412598
182 37.8486938476562
183 37.5516128540039
184 37.2772445678711
185 36.7542877197266
186 37.0196800231934
187 36.8709487915039
188 36.5793991088867
189 37.0893707275391
190 37.9901733398438
191 37.9926567077637
192 37.0303726196289
193 33.8661041259766
194 34.153263092041
195 35.5441818237305
196 27.5486392974854
197 22.9590816497803
198 26.5189552307129
199 24.8286609649658
};
\end{axis}

\end{tikzpicture}

%\caption{}
\label{fig:more_td_ewc}   
\end{subfigure}
     \caption{Performance comparison between \emph{One-step fine-tuning} and fine-tuning baselines in terms of Actual v.s. Predicted wind power values of target domain WT8.}
    \label{fig:more_avsp_td}
\end{figure*}
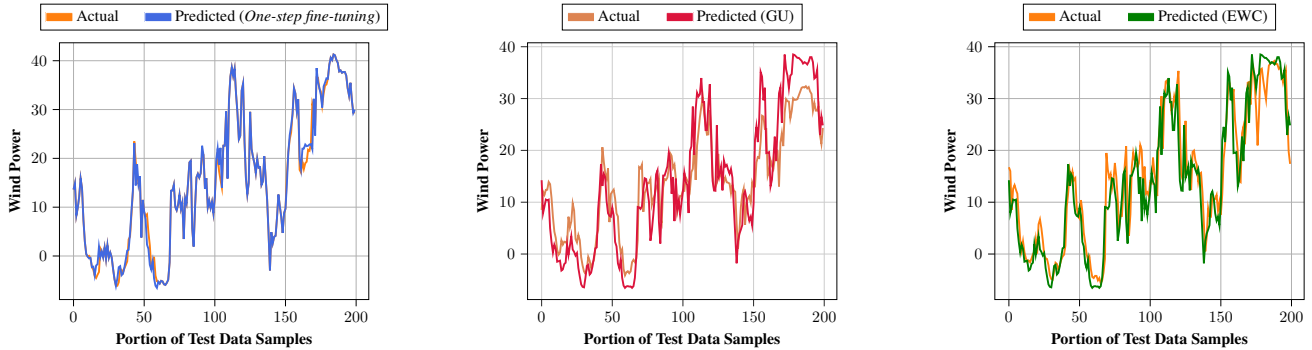

%% file: cf.tex
\begin{figure}[t!]
    \centering
    % This file was created with tikzplotlib v0.10.1.
\begin{tikzpicture}[scale=0.5]
\pgfplotsset{
every axis legend/.append style={at={(1,1.1)}, anchor=center,legend columns = 3},
legend style={/tikz/every even column/.append style={column sep=0.5cm}}}

\definecolor{crimson}{RGB}{220,20,60}
\definecolor{darkgray176}{RGB}{176,176,176}
\definecolor{green}{RGB}{0,128,0}
\definecolor{royalblue}{RGB}{65,105,225}

\begin{groupplot}[group style={group size=2 by 1, vertical sep =20pt, horizontal sep=70pt}]
\nextgroupplot[
tick align=outside,
tick pos=left,
x grid style={darkgray176},
xlabel={\textbf{Heat pump Deployment Sites (Target Domains)}},
xmin=-0.3625, xmax=4.8625,
xmajorgrids,
xtick style={color=black},
xtick={0.25,1.25,2.25,3.25,4.25},
xticklabels={S5,S8,S15,S49,NIST},
y grid style={darkgray176},
ylabel={\textbf{RMSE in S4}},
ymajorgrids,
ymin=0, ymax=0.9303,
ytick style={color=black}
]
\addlegendimage{area legend,pattern=north west lines, color=crimson,draw=black}
\addlegendimage{area legend,pattern=north west lines, color=green,draw=black}
\addlegendimage{area legend,pattern=north west lines, color=royalblue,draw=black}
\addlegendentry{GU}
\addlegendentry{EWC}
\addlegendentry{\emph{One-step fine-tuning}}

\draw[draw=black,fill=crimson] (axis cs:-0.125,0) rectangle (axis cs:0.125,0.347);
\draw[draw=black,fill=crimson] (axis cs:0.875,0) rectangle (axis cs:1.125,0.501);
\draw[draw=black,fill=crimson] (axis cs:1.875,0) rectangle (axis cs:2.125,0.886);
\draw[draw=black,fill=crimson] (axis cs:2.875,0) rectangle (axis cs:3.125,0.385);
\draw[draw=black,fill=crimson] (axis cs:3.875,0) rectangle (axis cs:4.125,0.34);
\draw[draw=black,fill=green] (axis cs:0.125,0) rectangle (axis cs:0.375,0.295);
\draw[draw=black,fill=green] (axis cs:1.125,0) rectangle (axis cs:1.375,0.419);
\draw[draw=black,fill=green] (axis cs:2.125,0) rectangle (axis cs:2.375,0.86);
\draw[draw=black,fill=green] (axis cs:3.125,0) rectangle (axis cs:3.375,0.246);
\draw[draw=black,fill=green] (axis cs:4.125,0) rectangle (axis cs:4.375,0.279);
\draw[draw=black,fill=royalblue] (axis cs:0.375,0) rectangle (axis cs:0.625,0.224);
\draw[draw=black,fill=royalblue] (axis cs:1.375,0) rectangle (axis cs:1.625,0.312);
\draw[draw=black,fill=royalblue] (axis cs:2.375,0) rectangle (axis cs:2.625,0.227);
\draw[draw=black,fill=royalblue] (axis cs:3.375,0) rectangle (axis cs:3.625,0.223);
\draw[draw=black,fill=royalblue] (axis cs:4.375,0) rectangle (axis cs:4.625,0.226);

\nextgroupplot[
tick align=outside,
tick pos=left,
x grid style={darkgray176},
xlabel={\textbf{Wind Turbines (Target Domains)}},
xmajorgrids,
xmin=-0.3625, xmax=4.8625,
xtick style={color=black},
xtick={0.25,1.25,2.25,3.25,4.25},
xticklabels={WT4,WT6,WT7,WT8,WT9},
y grid style={darkgray176},
ylabel={\textbf{RMSE in WT11}},
ymajorgrids,
ymin=0, ymax=0.15645,
ytick style={color=black}
]
\draw[draw=black,fill=crimson] (axis cs:-0.125,0) rectangle (axis cs:0.125,0.122);
\draw[draw=black,fill=crimson] (axis cs:0.875,0) rectangle (axis cs:1.125,0.133);
\draw[draw=black,fill=crimson] (axis cs:1.875,0) rectangle (axis cs:2.125,0.149);
\draw[draw=black,fill=crimson] (axis cs:2.875,0) rectangle (axis cs:3.125,0.129);
\draw[draw=black,fill=crimson] (axis cs:3.875,0) rectangle (axis cs:4.125,0.136);
\draw[draw=black,fill=green] (axis cs:0.125,0) rectangle (axis cs:0.375,0.125);
\draw[draw=black,fill=green] (axis cs:1.125,0) rectangle (axis cs:1.375,0.122);
\draw[draw=black,fill=green] (axis cs:2.125,0) rectangle (axis cs:2.375,0.14);
\draw[draw=black,fill=green] (axis cs:3.125,0) rectangle (axis cs:3.375,0.127);
\draw[draw=black,fill=green] (axis cs:4.125,0) rectangle (axis cs:4.375,0.134);
\draw[draw=black,fill=royalblue] (axis cs:0.375,0) rectangle (axis cs:0.625,0.119);
\draw[draw=black,fill=royalblue] (axis cs:1.375,0) rectangle (axis cs:1.625,0.117);
\draw[draw=black,fill=royalblue] (axis cs:2.375,0) rectangle (axis cs:2.625,0.13);
\draw[draw=black,fill=royalblue] (axis cs:3.375,0) rectangle (axis cs:3.625,0.115);
\draw[draw=black,fill=royalblue] (axis cs:4.375,0) rectangle (axis cs:4.625,0.12);
\end{groupplot}

\end{tikzpicture}

    \caption{Catastrophic forgetting check on Source Domains S4 and WT11 using fine-tuned target models.}
    \label{fig:cf}
\end{figure}
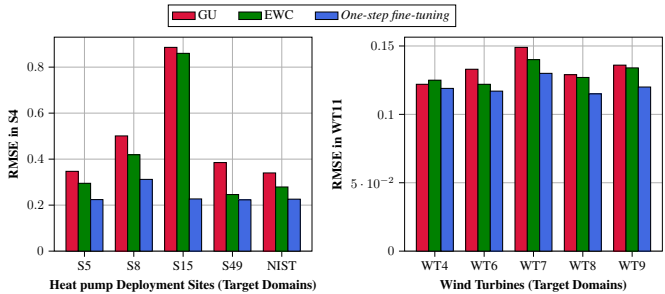

%% file: vp.tex
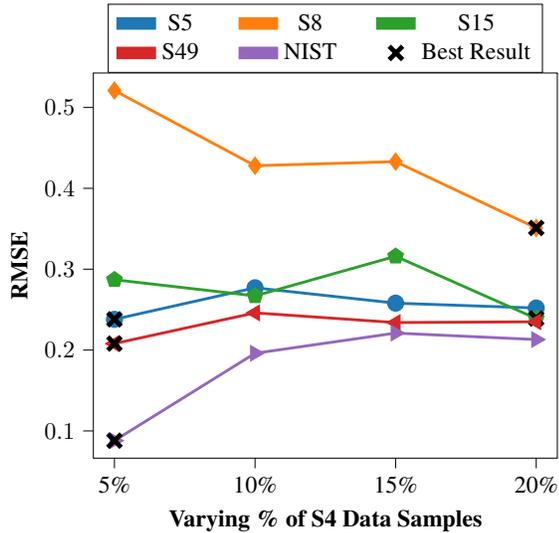
\begin{figure}[t!]
    \centering
 % This file was created with tikzplotlib v0.10.1.
\begin{tikzpicture}[scale=0.9]
\pgfplotsset{
every axis legend/.append style={at={(0.5, 1.09)}, anchor=center,legend columns = 3},
legend style={/tikz/every even column/.append style={column sep=0.5cm}}}

\definecolor{crimson2143940}{RGB}{214,39,40}
\definecolor{darkgray176}{RGB}{176,176,176}
\definecolor{darkorange25512714}{RGB}{255,127,14}
\definecolor{forestgreen4416044}{RGB}{44,160,44}
\definecolor{mediumpurple148103189}{RGB}{148,103,189}
\definecolor{steelblue31119180}{RGB}{31,119,180}

\begin{axis}[
tick align=outside,
tick pos=left,
x grid style={darkgray176},
xlabel={\textbf{Varying \% of S4 Data Samples}},
xmin=-0.15, xmax=3.15,
xtick style={color=black},
xtick={0,1,2,3},
xticklabels={5\%,10\%,15\%,20\%},
y grid style={darkgray176},
ylabel={\textbf{RMSE}},
ymin=0.06635, ymax=0.54265,
ytick style={color=black}
]
\addlegendimage{area legend,pattern=north west lines, color=steelblue31119180,draw=white}
\addlegendimage{area legend,pattern=north west lines, color=darkorange25512714,draw=white}
\addlegendimage{area legend,pattern=north west lines, color=forestgreen4416044,draw=white}
\addlegendimage{area legend,pattern=north west lines, color=crimson2143940,draw=white}
\addlegendimage{area legend,pattern=north west lines, color=mediumpurple148103189,draw=white}
\addlegendimage{only marks, draw=black, mark=x, mark size=4, mark options={line width=2pt}}
\addlegendentry{S5}
\addlegendentry{S8}
\addlegendentry{S15}
\addlegendentry{S49}
\addlegendentry{NIST}
\addlegendentry{Best Result}

\addplot [very thick, steelblue31119180, mark=*, mark size=3, mark options={solid}]
    table {%
        0 0.238
        1 0.277
        2 0.258
        3 0.252
    };

\addplot [draw=black, mark=x, mark size=4, mark options={line width=2pt}] coordinates {(0, 0.238)};

\addplot [very thick, darkorange25512714, mark=diamond*, mark size=3, mark options={solid}]
    table {%
        0 0.521
        1 0.428
        2 0.433
        3 0.351
    };

\addplot [draw=black, mark=x, mark size=4, mark options={line width=2pt}] coordinates {(3, 0.351)};

\addplot [very thick, forestgreen4416044, mark=pentagon*, mark size=3, mark options={solid}]
    table {%
        0 0.287
        1 0.267
        2 0.316
        3 0.239
    };

\addplot [draw=black, mark=x, mark size=4, mark options={line width=2pt}] coordinates {(3, 0.239)};

\addplot [very thick, crimson2143940, mark=triangle*, mark size=3, mark options={solid,rotate=90}]
    table {%
        0 0.208
        1 0.246
        2 0.234
        3 0.235
    };

\addplot [draw=black, mark=x, mark size=4, mark options={line width=2pt}] coordinates {(0, 0.208)};

\addplot [very thick, mediumpurple148103189, mark=triangle*, mark size=3, mark options={solid,rotate=270}]
    table {%
        0 0.088
        1 0.196
        2 0.221
        3 0.213
    };
    
\addplot [draw=black, mark=x, mark size=4, mark options={line width=2pt}] coordinates {(0, 0.088)};

\end{axis}
\end{tikzpicture}
    \caption{Performance of \emph{One-step fine-tuning} when varying the percentage of data samples from source domain S4.}
    \label{fig:varyingpercent}
\end{figure}

%% file: ft.tex
\begin{figure}[t!]
    \centering
    % This file was created with tikzplotlib v0.10.1.
\begin{tikzpicture}[scale=0.9]
\pgfplotsset{
every axis legend/.append style={at={(0.5, 1.069)}, anchor=center,legend columns = 3},
legend style={/tikz/every even column/.append style={column sep=0.1cm}}}

\definecolor{darkgray176}{RGB}{176,176,176}
\definecolor{olive}{RGB}{128,128,0}
\definecolor{royalblue}{RGB}{65,105,225}
\definecolor{saddlebrown}{RGB}{139,69,19}

\begin{axis}[
tick align=outside,
tick pos=left,
x grid style={darkgray176},
xlabel={\textbf{Heat pump Deployment Sites (Target Domains)}},
xmin=-0.3625, xmax=4.8625,
xmajorgrids,
xtick style={color=black},
xtick={0.25,1.25,2.25,3.25,4.25},
xticklabels={S5,S8,S15,S49,NIST},
y grid style={darkgray176},
ymajorgrids,
ylabel={\textbf{RMSE}},
ymin=0, ymax=0.61425,
ytick style={color=black}
]
\addlegendimage{area legend,pattern=north west lines, color=saddlebrown,draw=white}
\addlegendimage{area legend,pattern=north west lines, color=olive,draw=white}
\addlegendimage{area legend,pattern=north west lines, color=royalblue,draw=white}
\addlegendentry{Top layer unfreeze}
\addlegendentry{w/o GU}
\addlegendentry{Ours}

\draw[draw=black,fill=saddlebrown] (axis cs:-0.125,0) rectangle (axis cs:0.125,0.331);
\draw[draw=black,fill=saddlebrown] (axis cs:0.875,0) rectangle (axis cs:1.125,0.585);
\draw[draw=black,fill=saddlebrown] (axis cs:1.875,0) rectangle (axis cs:2.125,0.346);
\draw[draw=black,fill=saddlebrown] (axis cs:2.875,0) rectangle (axis cs:3.125,0.241);
\draw[draw=black,fill=saddlebrown] (axis cs:3.875,0) rectangle (axis cs:4.125,0.169);
\draw[draw=black,fill=olive] (axis cs:0.125,0) rectangle (axis cs:0.375,0.252);
\draw[draw=black,fill=olive] (axis cs:1.125,0) rectangle (axis cs:1.375,0.481);
\draw[draw=black,fill=olive] (axis cs:2.125,0) rectangle (axis cs:2.375,0.296);
\draw[draw=black,fill=olive] (axis cs:3.125,0) rectangle (axis cs:3.375,0.226);
\draw[draw=black,fill=olive] (axis cs:4.125,0) rectangle (axis cs:4.375,0.161);
\draw[draw=black,fill=royalblue] (axis cs:0.375,0) rectangle (axis cs:0.625,0.238);
\draw[draw=black,fill=royalblue] (axis cs:1.375,0) rectangle (axis cs:1.625,0.351);
\draw[draw=black,fill=royalblue] (axis cs:2.375,0) rectangle (axis cs:2.625,0.239);
\draw[draw=black,fill=royalblue] (axis cs:3.375,0) rectangle (axis cs:3.625,0.208);
\draw[draw=black,fill=royalblue] (axis cs:4.375,0) rectangle (axis cs:4.625,0.088);
\end{axis}

\end{tikzpicture}

    \caption{Performance of \emph{One-step fine-tuning} compared with different freezing techniques.}
    \label{fig:freezing_tech}
\end{figure}
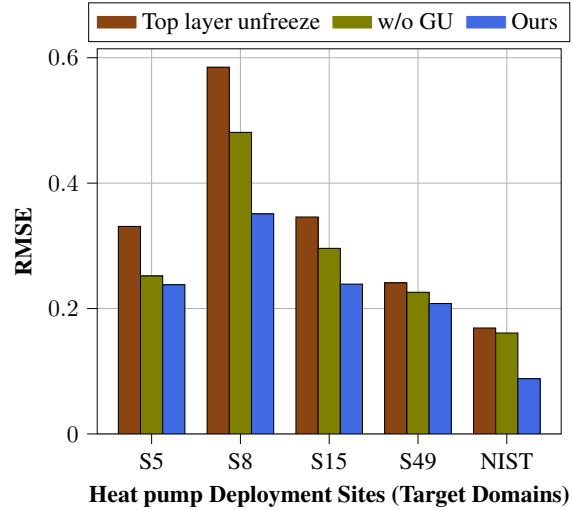